\documentclass[10pt,twocolumn,letterpaper]{article}
\usepackage[pagenumbers]{cvpr}
\definecolor{cvprblue}{rgb}{0.21,0.49,0.74}
\usepackage[pagebackref,breaklinks,colorlinks,allcolors=cvprblue]{hyperref}

\usepackage{url}
\usepackage{graphicx}
\usepackage{booktabs}
\usepackage{multirow}
\usepackage{makecell}
\usepackage{xcolor}
\usepackage{placeins}
\usepackage[table]{xcolor}

\def\paperID{} 
\def\confName{CVPR}
\def\confYear{2026}

\title{Rethinking VLMs for Image Forgery Detection and Localization}

\author{
Shaofeng Guo \qquad Jiequan Cui\thanks{Corresponding Author (\href{jiequancui@gmail.com}{jiequancui@gmail.com}).} \qquad Richang Hong \\
Hefei University of Technology \\
Key Laboratory of Knowledge Engineering with Big Data, 
Ministry of Education \\
{\tt\small shaofengGuo@mail.hfut.edu.cn, \{jiequancui, hongrc.hfut\}@gmail.com} \\
}
\begin{document}
\maketitle

\begin{abstract}
     With the rapid rise of Artificial Intelligence Generated Content (AIGC), image manipulation has become increasingly accessible, posing significant challenges for image forgery detection and localization (IFDL).
     In this paper, we study how to fully leverage vision-language models (VLMs) to assist the IFDL task.
     In particular, we observe that priors from VLMs hardly benefit the detection and localization performance and even have negative effects due to their inherent biases toward semantic plausibility rather than authenticity.
     Additionally, the location masks explicitly encode the forgery concepts, which can serve as extra priors for VLMs to ease their training optimization, thus enhancing the interpretability of detection and localization results.
     Building on these findings, we propose a new IFDL pipeline --- \textit{IFDL-VLM}.
     To demonstrate the effectiveness of our method, we conduct experiments on 9 popular benchmarks and assess the model performance under both settings of in-domain and cross-dataset generalization. The experimental results show that we consistently achieve new state-of-the-art performance in detection, localization, and interpretability. Code is available at \href{https://github.com/sha0fengGuo/IFDL-VLM}{https://github.com/sha0fengGuo/IFDL-VLM}.
\end{abstract}
\section{Introduction}

Security and robustness have long been a critical research topic in machine learning and computer vision~\cite{madry2017towards,he2022masked, menon2020long,cui2021parametric}. Adversarial robustness~\citep{madry2017towards, zhang2019theoretically, cui2021learnable, wang2023better, cui2024decoupled, cui2025generalized}, out-of-distribution (OOD) generalization~\citep{he2022masked, cui2023generalized}, OOD detection~\citep{wei2022mitigating, liu2020energy, liang2017enhancing, sun2022out, liu2024typicalness}, and distribution shift~\cite{wiles2021fine, cui2022reslt, cui2024classes, cui2025generative} aim to enhance the reliability of model predictions. In contrast, data security poses another growing challenge~\citep{guillaro2023trufor, guo2023hierarchical, wang2022objectformer, dong2022mvss, kwon2021cat, zhang2024editguard}. Maliciously tampered or synthetic images can mislead public opinion and distort historical records. In the era of Artificial Intelligence Generated Content (AIGC), advanced image editing algorithms~\citep{ruiz2023dreambooth, gal2022image, mokady2023null, hertz2023delta, song2024doubly} can already produce highly realistic or create content that resembles real-world events, further amplifying the risks associated with data authenticity and security.

\begin{figure*}
    \centering
    \includegraphics[width=1.0\linewidth]{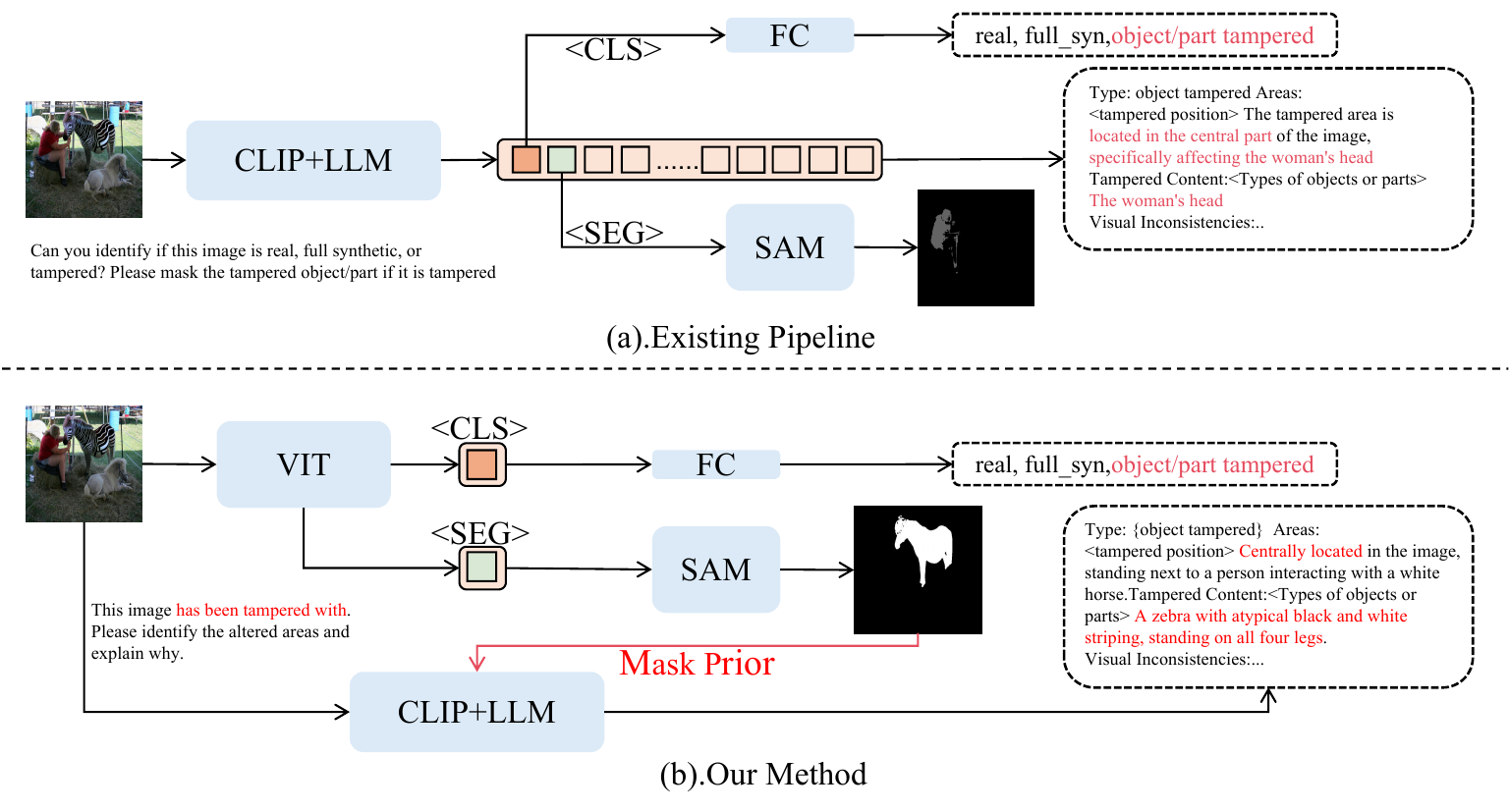}
    \vspace{-0.3in}
    \caption{Comparison with the existing IFDL pipeline. (a) Existing IFDL algorithms enhance the interpretability with language explanations from VLMs. Meanwhile, the $<SEG>$ token and the $<CLS>$ token for localization and detection are derived from the responses of VLMs. (b) Our IFDL method decouples the optimization of language explanations and detection\&localization. Moreover, the detection\&localization results are used as extra inputs to assist the optimization of language explanation with VLMs. }
    \label{fig:comparsion_with_prior_work}
    \vspace{-0.1in}
\end{figure*}

The main challenges of image forgery detection and localization (IFDL) are threefold: (I) jointly leveraging high-level semantics and low-level artifacts, (II) generalizing across diverse forgery types, and (III) ensuring explainability and trustworthiness of detection and localization results. Prior work~\citep{dong2022mvss, ma2023iml} introduces edge-detection regularization to guide models toward low-level cues such as boundary inconsistencies and lighting mismatches. Another line of research improves generalization by ensembling multiple forgery experts~\citep{yu2025reinforced} or collecting large-scale datasets covering diverse manipulations~\citep{qu2024towards,novozamsky2020imd2020}. Recently, vision-language models (VLMs) have been adopted to enhance interpretability and trustworthiness. In this paper, we revisit the role of VLMs and explore how to fully leverage their capabilities to improve IFDL performance.

\noindent{\bf Our Motivation.}
As shown in Figure~\ref{fig:comparsion_with_prior_work}(a), prior works such as SIDA and FakeShield integrate CLIP~\citep{radford2021learning}, large language models (LLMs)~\citep{vicuna2023}, and SAM~\citep{kirillov2023segment} into a unified pipeline with 3 steps: (1) carefully designed text prompts and image features from CLIP are fed into the LLM; (2) the LLM generates natural language explanations for detection and localization; and (3) special tokens $<CLS>$ and $<SEG>$ in the LLM’s response trigger the linear classifier and SAM to produce forged types and masks highlighting the forged regions.

While the vision-language alignment priors from CLIP models and the language knowledge from LLMs have proven beneficial for downstream tasks such as image captioning and VQA~\citep{liu2023visual, zhu2023minigpt}, \textit{it remains unclear whether these priors can effectively enhance forgery detection and localization, given that the models are not exposed to forgery-related concepts during their pretraining on large-scale data}.
Additionally, we also explore the reverse perspective: \textit{can detection and localization results be leveraged to improve the optimization process of VLMs and further enhance their interpretability?}

\noindent{\bf Our Method.}
To answer the above two questions, we propose a new framework, \textit{IFDL-VLM}, as illustrated in Figure~\ref{fig:comparsion_with_prior_work}(b). The framework consists of two training stages:
\begin{itemize}
\item \textit{Stage-1:} Jointly train a ViT encoder with SAM to perform forgery detection and localization.
\item \textit{Stage-2:} Leverage the derived detection and localization results in \textit{Stage-1} as additional inputs to fine-tune the VLM. Note that the detection and localization results explicitly define the forgery concepts. Then the VLM no longer needs to infer them from data, thereby easing the optimization and enhancing the interpretability.
\end{itemize}

Unlike prior work~\citep{huang2025sida, xu2024fakeshield}, our \textit{IFDL-VLM} framework decouples the optimization of detection\&localization from the generation of VLM-based language explanations and further leverages the derived localization masks to promote the VLM optimization. 
Empirically, we find that \textit{localization masks from \textit{Stage-1} can effectively serve as auxiliary inputs of VLMs to improve the interpretability of the detection and localization results, whereas the priors from VLMs contribute little to detection and localization performance}.

\noindent{\bf Experimental Results.}
To validate the effectiveness of our approach, we evaluate it on 9 popular benchmarks under both in-domain and cross-dataset generalization settings. Our method achieves new state-of-the-art performance across detection, localization, and interpretability. For localization, we obtain 0.65 IoU on SID-Set and an average IoU of 0.47 on 8 datasets (CASIA1+, IMD2020, Columbia, NIST, DSO, Korus, DeepFake, and AIGC Editing), surpassing the SIDA by \textbf{21\%} IoU and outperforming FakeShield by \textbf{13\%} IoU respectively. Additionally, in GPT-5 automated evaluation of interpretability, our method achieves an overall score of 2.44, representing \textbf{0.77} improvement over SIDA.

Our contributions are summarized as follows:
\begin{itemize}
    \item We propose the IFDL-VLM framework to study how to fully exploit VLMs for the IFDL task.
    \item With the IFDL-VLM method, we find that taking localization masks as auxiliary inputs for VLMs can benefit their training optimization, thus enhancing the interpretability. Meanwhile, the inherent VLM biases towards semantic plausibility rather than authenticity can hinder the IFDL task performance.
    \item Extensive experiments are conducted on 9 popular datasets. Notably, our IFDL-VLM method achieves new state-of-the-art performance across detection, localization, and interpretability.
\end{itemize}

\section{Related Work}

\noindent\textbf{AIGC and the IFDL Challenge.}
Diffusion models~\citep{ho2020denoising,rombach2022high}, GANs~\citep{karras2019style}, and autoregressive transformers~\citep{vaswani2017attention} enable photorealistic image synthesis, blurring the line between real and fake. Unlike traditional manipulations (splicing, copy-move), AIGC outputs lack low-level artifacts, instead exhibiting subtle semantic inconsistencies. Hybrid forgeries — AI-generated content blended with conventional edits (e.g., Photoshop) and post-processing (compression, upscaling) — further erase forensic traces, rendering legacy IFDL methods ineffective. This demands new datasets and detection paradigms for artifact-scarce, semantically plausible forgeries.

\noindent\textbf{Image Forgery Detection \& Localization (IFDL).}
IFDL methods primarily aim at forensic trace modeling — detecting and localizing high-level misaligned semantics and low-level inconsistencies (\textit{e.g.}, noise, edge, compression artifacts) introduced by image manipulation. 
Modern frameworks often integrate handcrafted priors, like SRM DCT, BayarConv, with learning-based components to adaptively fuse and enhance forensic signals — as exemplified by MVSS-Net~\citep{dong2022mvss} (BayarConv + multi-scale edge), CAT-Net~\citep{kwon2021cat} (DCT + RGB fusion), and DiffForensics~\citep{yu2024diffforensics} (diffusion pretraining + edge enhancement). 
Beyond trace modeling, two additional research directions have gained attention: (I) domain generalization, aiming to improve robustness to unseen manipulation types; and (II) enhancing the interpretability and trustworthiness of detection and localization results.

\noindent\textbf{Multimodal Large Language Models (MLLMs).}
Multi-modal models, like CLIP~\citep{radford2021learning}, align visual-textual representations via cross-modal pretraining, enabling the integration of vision and language analysis via combining it with large language models (LLMs).
These innovations have had a notable impact in domains requiring both visual understanding and reasoning, such as content generation, interactive dialogue, and multimodal reasoning tasks. 
By leveraging large-scale multimodal datasets, MLLMs are able to handle more complex challenges, such as image captioning~\cite{liu2023visual, li2023blip}, VQA~\cite{liu2023visual, li2023blip}, image segmentation \citep{lai2024lisa} and deepfake detection. Recent works have demonstrated the power of MLLMs in detecting and localizing image forgeries \citep{chang2023antifakeprompt,liu2024forgerygpt,xu2024fakeshield,huang2025sida}, combining image analysis with textual descriptions to create more robust frameworks for identifying subtle manipulations while enhancing interpretability.

\section{Method --- IFDL-VLM}


\subsection{Do Vision-Language Alignment Priors Benefit IFDL Performance?}
\label{sec:vlm_bias}
CLIP and LLMs have demonstrated strong performance on downstream tasks such as image captioning and VQA, benefiting from vision-language alignment and common-sense reasoning. However, these models lack priors for localization and low-level cues (e.g., edges), which are critical for image forgery detection~\citep{dong2022mvss, kwon2021cat}. 
Moreover, since VLMs are originally pretrained for global semantic alignment on natural images, even with fine-tuning on IFDL datasets, inherent biases can still hinder task performance.

\noindent{\bf Semantic Plausibility vs. Authenticity.}
The prevalent pipeline of VLMs leverages the \textit{frozen CLIP visual tokens} as inputs to finetune LLMs with corresponding language annotations. However, CLIP models are optimized to align the high-level visual semantics ( object or scene) with language concepts. As a result, a forged image in which certain objects or scene instances have been replaced, swapped, or manipulated may still produce visual token representations that remain well-aligned with the original textual annotation, making the visual token representations indistinguishable between authentic and manipulated images, thereby potentially hindering IFDL performance.

\begin{figure}[t]
    \centering
    \includegraphics[width=1.0\linewidth]{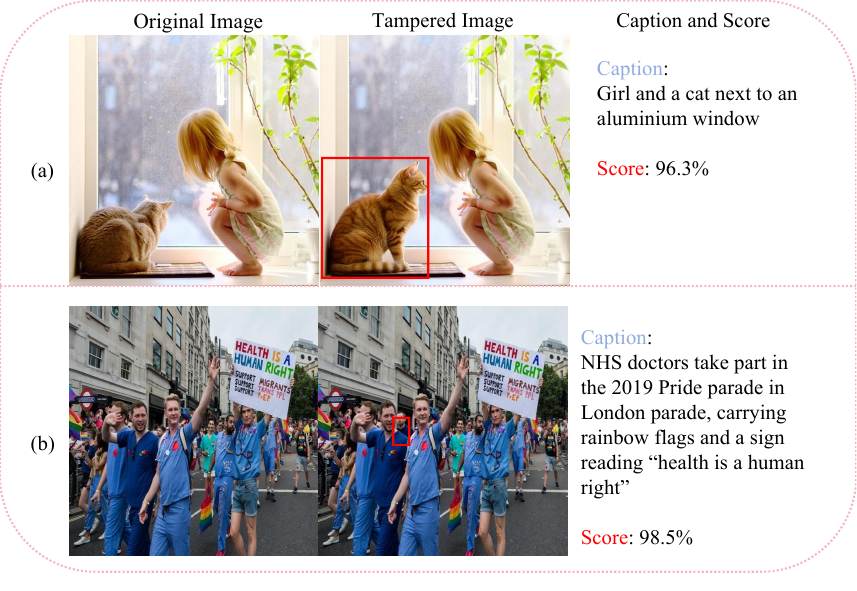}
    \vspace{-0.1in}
    \caption{
   VLMs exhibit a bias toward \textit{semantic plausibility} rather than \textit{authenticity}. As a result, they are often insensitive to semantically consistent or subtle forgeries, such as the replacement of a cat instance in (a) or the addition of a new person in (b). This inherent bias ultimately hinders IFDL performance.
    }
    \label{fig:vlm_bias_examples}
    \vspace{-0.1in}
\end{figure}

Vision-language models (VLMs) exhibit a strong preference for semantic plausibility over authenticity, which leads them to overlook semantically consistent and subtle forgeries.
As illustrated in Figure~\ref{fig:vlm_bias_examples}, despite cat instance changes in Figure~\ref{fig:vlm_bias_examples}(a) and a new person is added in Figure~\ref{fig:vlm_bias_examples}(b), their captions still reflect the semantics in the images, indicating that vision language alignment optimization can lead to representations insensitivity to image authenticity. 
Quantitatively, CLIP visual feature cosine similarities reach \textbf{96.3\%} and \textbf{98.5\%} for the cat and parade pairs, respectively, confirming that the representations between authentic and manipulated images are hard to distinguish. This reveals that VLMs prioritize high-level scene coherence over low-level visual authenticity, rendering them ineffective for fine-grained forgery detection. 

\begin{figure*}[t!]
    \centering
    \includegraphics[width=1.0\linewidth]{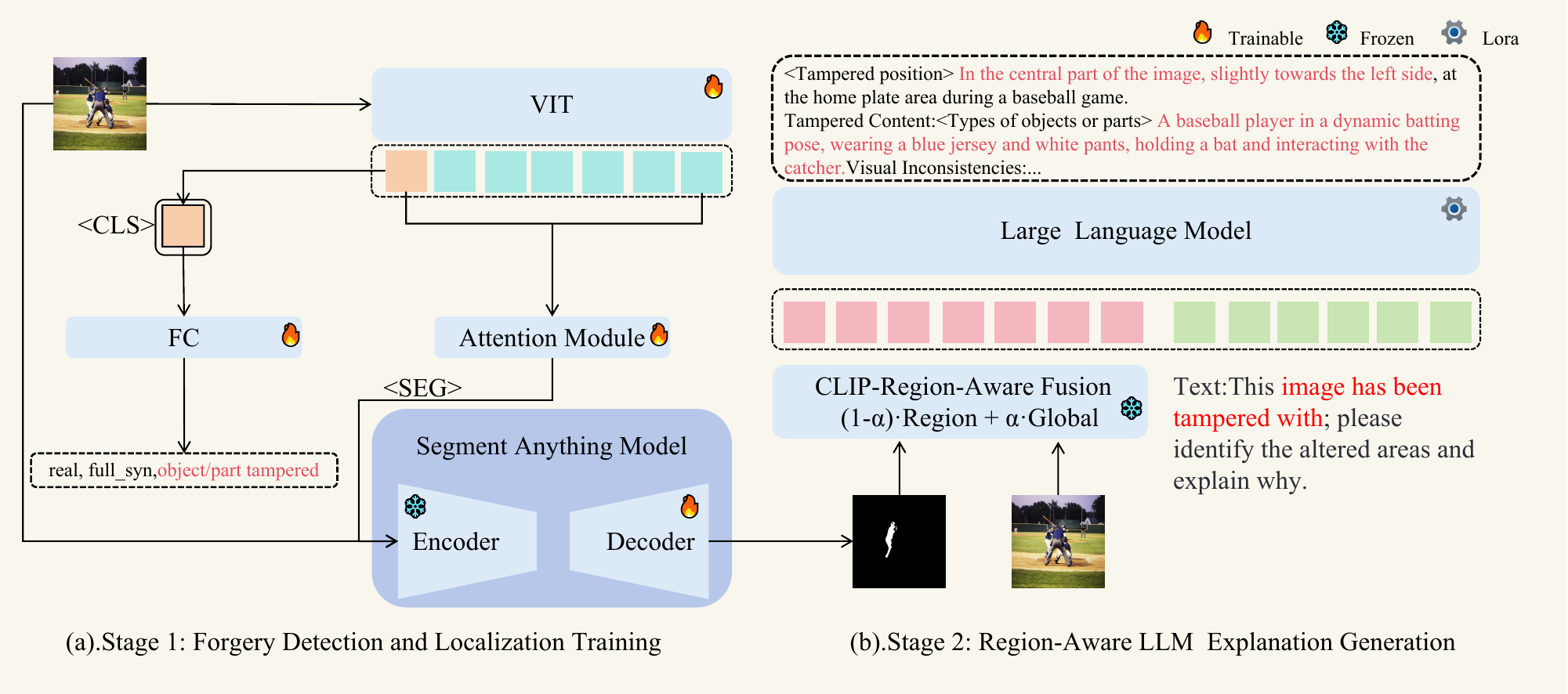}
    \caption{Our IFDL-VLM framework. IFDL-VLM method decouples the optimization of detection\&localization from the language explanation generation. (a) Detection\&localization expert training with VIT and SAM. (b) Region-aware visual feature enhancement technique for language explanation generation.}
    \label{fig:ifdl_vlm}
    \vspace{-0.15in}
\end{figure*}

\subsection{IFDL-VLM}
To address the problems in Section~\ref{sec:vlm_bias}, we propose the IFDL-VLM framework, as illustrated in Figure~\ref{fig:ifdl_vlm}. Our IFDL-VLM method features two key components: (1) decoupled optimization of detection\&localization and language explanation generation, and (2) leveraging detection\&localization results to enhance language interpretability. We describe the detailed implementations as follows.

\noindent{\bf Decoupled Optimization of Detection\&Localization and Language Explanation Generation.}
Instead of jointly optimizing detection, localization, and language explanations within a single VLM, we decouple the IFDL task into two stages:
(i) forgery detection and localization optimization using a trainable ViT backbone integrated with SAM, and
(ii) region-aware enhancement for language explanation generation via VLMs.

As discussed in Section~\ref{sec:vlm_bias}, VLMs lack priors on forgery-related concepts, and their inherent semantic plausibility biases can even hinder IFDL performance. 
To address this issue, we propose to learn an expert ViT model for detection and localization. Specifically, as shown in Figure~\ref{fig:ifdl_vlm}(a), in \textit{Stage-1}, the input image is processed by a learnable ViT backbone to produce both $<SEG>$ and $<CLS>$ tokens. The $<SEG>$ token is passed to SAM to generate localization masks, while the $<CLS>$ token is used for linear classification. Overall, the training objective of \textit{Stage-1} is as follows:
\begin{eqnarray}
    \mathcal{L}_{loc} \!&=&\! \lambda_{bce} \mathcal{L}_{bce}(\hat{M}, M) \!+\! \lambda_{dice} \mathcal{L}_{dice}(\hat{M}, M) \\
    \mathcal{L}_{det} \!&=&\! \lambda_{det} \mathcal{L}_{ce}(\hat{D}, D) \\
    \mathcal{L}_{st-1} \!&=&\! \mathcal{L}_{det} + \mathcal{L}_{loc}, 
\end{eqnarray}
where ($\lambda_{bce}$, $\lambda_{dice}$, $\lambda_{det}$) represents hyper-parameters and is set to (1.0, 1.0, 1.0) by default, $\hat{M}$ and $M$ are the predicted and ground-truth localization masks respectively, $\hat{D}$ and $D$ are the predicted and ground-truth forgery types individually, $\mathcal{L}_{bce}$ denotes binary cross-entropy loss while $\mathcal{L}_{dice}$ represents dice loss. 

\noindent{\bf Localization Masks Encode the Forgery-Related Concepts.}
The inherent semantic plausibility biases in Section~\ref{sec:vlm_bias} can also have negative effects on language interpretability due to the indistinguishable visual tokens between authentic and forged images. To mitigate the problem, we propose to incorporate the derived localization masks from \textit{Stage-1} to enhance the visual features. 
Specifically, as shown in Figure~\ref{fig:ifdl_vlm}(b), the region-aware visual feature enhancement technique is developed. Overall, the training objective of \textit{Stage-2} is as follows:
\begin{eqnarray}
    T_{vis} \!&=&\! \alpha CLIP(x) \!+\! (1 \!-\! \alpha) CLIP(x \odot M)\label{eq:vis_enhance}\\
    \hat{y}_{des} \!&=&\! LLM(T_{vis}, text) \\
    \mathcal{L}_{st-2} \!&=&\! \mathcal{L}_{ce}(\hat{y}_{des}, y_{des}),
\end{eqnarray}
where $x$ is an input image, $T_{vis}$ is the visual tokens, $\hat{y}_{des}$ represents the outputs of the LLM while $y_{des}$ denotes the ground-truth language annotation. 
Unless otherwise specified, we set $\alpha =0.5$. At inference, we replace the ground-truth mask $M$ with the predicted one $\hat{M}$ derived in \textit{Stage-1}.

On the one hand, Eq.~\eqref{eq:vis_enhance} enriches visual features with low-level cues in forged regions, improving the separability between authentic and manipulated image representations. On the other hand, since localization masks explicitly encode forgery-related concepts, LLMs are relieved from learning these concepts purely from data, thereby simplifying their training optimization. Thus, the quality of generated language explanations is enhanced.


\begin{table*}[htp!]
    \caption{Comparisons with previous methods regarding detection performance on SID-Set. Numbers are borrowed from SIDA~\cite{huang2025sida}. The best results are in bold. The overall accuracy and F1 are calculated as the average of the values from the three classification categories.}
    \vspace{-0.1in}
    \label{tab:detection}
    \centering
    \setlength{\tabcolsep}{4pt} 
    \renewcommand{\arraystretch}{1.2} 
    
    \begin{tabular}{l c c c c c c c c c}
        \toprule
        \multirow{2}{*}{Methods} & \multirow{2}{*}{Year} & 
        \multicolumn{2}{c}{Real} & 
        \multicolumn{2}{c}{Fully synthetic} & 
        \multicolumn{2}{c}{Tampered} & 
        \multicolumn{2}{c}{Overall} \\
        \cmidrule(r){3-4} \cmidrule(r){5-6} \cmidrule(r){7-8} \cmidrule(r){9-10}
        & & Acc & F1 & Acc & F1 & Acc & F1 & Acc & F1 \\
        \midrule
        AntifakePrompt & 2024  
        & 0.65 & 0.79 & 0.94 & 0.97 & 0.31 & 0.47 & 0.63 & 0.74 \\
        CnnSpott & 2021 
        & 0.80 & 0.89 & 0.40 & 0.57 & 0.07 & 0.13 & 0.42 & 0.53 \\
        FreDect & 2020  
        & 0.84 & 0.91 & 0.17 & 0.29 & 0.12 & 0.21 & 0.37 & 0.47 \\
        Fusing & 2022 
        & 0.85 & 0.92 & 0.34 & 0.51 & 0.03 & 0.05 & 0.41 & 0.49 \\
        Gram-Net & 2020 
        & 0.70 & 0.82 & 0.94 & 0.97 & 0.01 & 0.02 & 0.55 & 0.60 \\
        UnivFD & 2023
        & 0.68 & 0.67 & 0.62 & 0.88 & 0.64 & 0.85 & 0.65 & 0.80 \\
        LGrad & 2023 
        & 0.65 & 0.79 & 0.84 & 0.91 & 0.07 & 0.13 & 0.52 & 0.61 \\
        LNP & 2023
        & 0.71 & 0.83 & 0.92 & 0.96 & 0.03 & 0.06 & 0.55 & 0.62 \\
        \midrule
        SIDA-7B & 2024 
        & 0.89 & \underline{0.91} & \underline{0.99} & \underline{0.99} & \underline{0.93} & \underline{0.91} & \underline{0.94} & \underline{0.94} \\
        SIDA-13B & 2024 
        & \underline{0.90} & \underline{0.91} & \underline{0.99} & \underline{0.99} & \underline{0.93} & \underline{0.91} & \underline{0.94} & \underline{0.94} \\
        \midrule
        \textbf{IFDL-VLM (Ours)} & - 
        & \textbf{0.998} & \textbf{0.996} & \textbf{1.000} & \textbf{1.000} & \textbf{0.994} & \textbf{0.996} & \textbf{0.997} & \textbf{0.998} \\
        \bottomrule
    \end{tabular}
    \vspace{-0.1in}
\end{table*}

\begin{table}[ht]
\centering
\caption{Comparison with previous methods regarding localization results on SID-Set.}
\vspace{-0.2in}
\label{tab:localization_results}
\vspace{5pt}
\renewcommand{\arraystretch}{1.15}
\begin{tabular}{c|c|ccc}
\toprule
Method & Year & AUC & F1 & IoU \\
\midrule
MVSS-Net & 2023 & 0.49 & 0.32 & 0.24 \\
HiFi-Net & 2023 & 0.64 & 0.46 & 0.21 \\
PSCC-Net & 2022 & 0.82 & 0.71 & 0.36 \\
LISA-7B-v1 & 2024 & 0.78 & 0.69 & 0.33 \\
SIDA-7B & 2024 & \underline{0.87} & \underline{0.74} & \underline{0.44} \\
\midrule
\textbf{IFDL-VLM (Ours)} & - & \textbf{0.99} & \textbf{0.87} & \textbf{0.65} \\
\bottomrule
\end{tabular}
\vspace{-0.1in}
\end{table}

\begin{figure*}[htp!]
  \centering
  \includegraphics[width=0.95\textwidth]{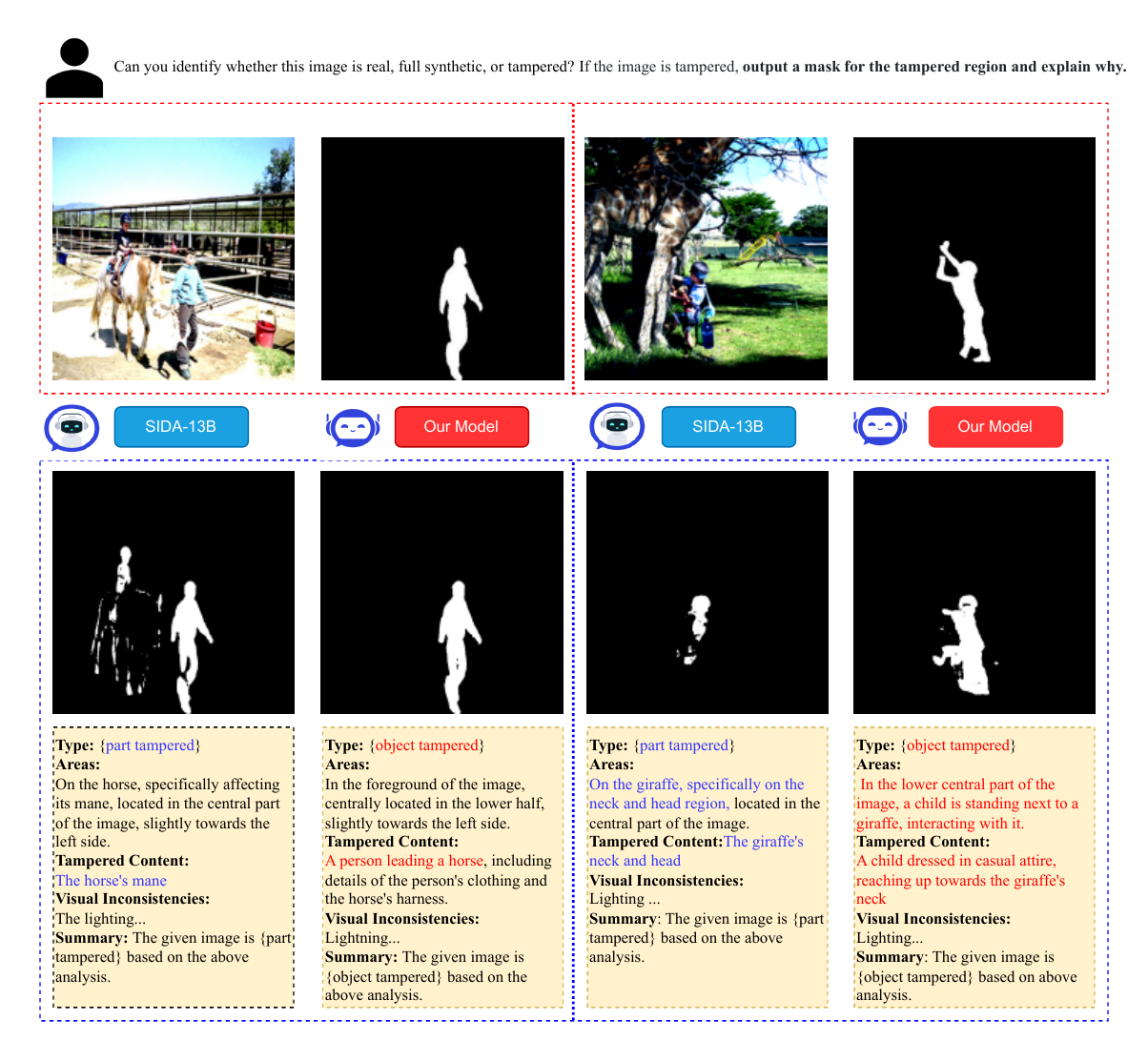}
  \vspace{-0.25in}
  \caption{Qualitative comparison on localization mask and language explanations.}
  \label{fig:visual_comparison}
  \vspace{-0.1in}
\end{figure*}

\section{Experiments}
\label{sec:Experiments}
\subsection{Experimental Setup}
\noindent\textbf{Datasets.}
To assess the effectiveness of our method, we follow two evaluation protocols:
\begin{itemize}
\item \textit{(i) SID-Set}. We perform experiments on SID-Set with the default 7:1:2 split, using three-way image-level labels (\emph{Real}, \emph{Fully Synthetic}, \emph{Tampered}) and binary pixel-level masks. Its description annotations (SID-Set-description) are also adopted as text supervision for Stage-2.
\item \textit{(ii) Cross-dataset generalization.} A cross-dataset setting where models are trained on CASIAv2, FantasticReality, DeepFake, and AIGC Editing datasets, and evaluated on 8 benchmarks (CASIA1+, IMD2020, Columbia, NIST, DSO, Korus, DeepFake, and AIGC Editing).
\end{itemize}

\noindent\textbf{Evaluation Metrics.}
For detection, we report image-level accuracy (ACC) and F1 scores. For localization, we provide Intersection over Union (IoU) and F1 scores. 
For both detection and localization, a default threshold of 0.5 is applied unless otherwise specified.
To evaluate interpretability, we perform automated evaluation with the GPT-5 model. Additionally, we also use Cosine Semantic Similarity (CSS) to assess the similarity between the predicted text and ground truth text by calculating the cosine similarity between their high-dimensional semantic vectors. Specifically, we respectively report CSS in 5 sections including \textit{Type, Areas, Tampered Content, Visual Inconsistencies, and Summary}. To emphasize ``where'', ``what'', and ``why'' in the language explanations, we aggregate the section scores with weights $\{0.05,0.35,0.40,0.15,0.05\}$.

\noindent\textbf{Implementation Details.}
We initialize the ViT backbone with CLIP-ViT-L/14. Given an input image, the global $<CLS>$ token is used for three-way classification, while patch-level features are projected and fused through multi-head attention to derive a $<SEG>$ token that serves as the prompt embedding for SAM. For localization, we freeze the SAM-H image encoder to extract high-resolution features and fine-tune its mask decoder. Input images are resized to $1024{\times}1024$ for SAM. For the language explanation generation in Stage-2, we adopt Vicuna-13B as our Large Language Model backbone, following the configuration of SIDA.

\subsection{Detection\&Localization Comparisons}
\label{sec:comparison_ifdl}
We structure our evaluation into two tracks: (i) experiments on SID-Set under the same setting as SIDA, and (ii) extended cross-dataset evaluations following FakeShield to assess the generalization ability of our models.

\subsubsection{Evaluation on SID-Set}
\label{subsec:sida_setting}

\noindent{\bf Training Setup.}
For fair comparisons, we follow SIDA to train and evaluate our models on SID-Set with the same settings. 
Specifically, we optimize the model using AdamW optimizer with an initial learning rate of $1e-5$, $\beta=(0.9,0.95)$, and no weight decay. 
A linear warmup–decay learning rate schedule is adopted: the learning rate is linearly increased from 0 to 1e-5 over the first 100 warmup steps, then decays over the remaining 9900 steps.
We use a batch size of 4 with gradient accumulation of 10. Mixed-precision training (FP16/BF16) is adopted, and the gradient clipping hyperparameter is set to 1.0.
Both detection (classification) and localization results are reported.

\noindent{ \bf Detection Accuracy and F1 scores.} 
Table~\ref{tab:detection} reports the detection performance against SIDA and other baselines. Our method, IFDL-VLM, achieves the highest accuracy and F1 score across all categories, with an overall $\textbf{99.7}\%$ Accuracy and $\textbf{99.8}\%$ F1 score respectively, substantially outperforming prior approaches. 

\begin{table*}[htp!]
\centering
\caption{Comparative results of tamper localization capabilities between competing IFDL methods and our IFDL-VLM, tested on MMTD-Set (Photoshop, DeepFake, AIGC-Editing). 
All numerical results for baseline methods are borrowed from FakeShield~\cite{xu2024fakeshield}. 
Results marked with an asterisk (*) are reproduced using the official checkpoint released by the authors.}
\vspace{-0.1in}
\label{FS_location_result_table}
\small
\setlength{\tabcolsep}{4pt} 
\renewcommand{\arraystretch}{1.2} 
\resizebox{\linewidth}{!}
{
\begin{tabular}{l *{18}{c}}
\toprule
\multirow{2}{*}{Method} 
& \multicolumn{2}{c}{CASIA1+} 
& \multicolumn{2}{c}{IMD2020} 
& \multicolumn{2}{c}{Columbia} 
& \multicolumn{2}{c}{NIST} 
& \multicolumn{2}{c}{DSO} 
& \multicolumn{2}{c}{Korus} 
& \multicolumn{2}{c}{DeepFake} 
& \multicolumn{2}{c}{AIGC-Editing}
& \multicolumn{2}{c}{Avg} \\
\cmidrule(r){2-3} \cmidrule(r){4-5} \cmidrule(r){6-7} \cmidrule(r){8-9} \cmidrule(r){10-11} \cmidrule(r){12-13} 
\cmidrule(r){14-15} \cmidrule(r){16-17} \cmidrule(r){18-19}
& IoU & F1 & IoU & F1 & IoU & F1 & IoU & F1 & IoU & F1 & IoU & F1 & IoU & F1 & IoU & F1 & IoU & F1 \\
\midrule
SPAN & 0.11 & 0.14 & 0.09 & 0.14 & 0.14 & 0.20 & 0.16 & 0.21 & 0.14 & 0.24 & 0.06 & 0.10 & 0.04 & 0.06 & 0.09 & 0.12 & 0.10 & 0.15 \\
ManTraNet & 0.09 & 0.13 & 0.10 & 0.16 & 0.04 & 0.07 & 0.14 & 0.20 & 0.08 & 0.13 & 0.02 & 0.05 & 0.03 & 0.05 & 0.07 & 0.12 & 0.07 & 0.11 \\
OSN & 0.47 & 0.51 & 0.38 & 0.47 & 0.58 & 0.69 & 0.25 & 0.33 & 0.32 & 0.45 & 0.14 & 0.19 & 0.11 & 0.13 & 0.07 & 0.09 & 0.29 & 0.37 \\
HiFi-Net & 0.13 & 0.18 & 0.09 & 0.14 & 0.06 & 0.11 & 0.09 & 0.13 & 0.18 & 0.29 & 0.01 & 0.06 & 0.07 & 0.11 & 0.13 & 0.22 & 0.10 & 0.15 \\
PSCC-Net & 0.36 & 0.46 & 0.22 & 0.32 & 0.64 & 0.74 & 0.18 & 0.26 & 0.22 & 0.33 & 0.15 & \underline{0.22} & 0.12 & 0.18 & 0.10 & 0.15 & 0.24 & 0.33 \\
CAT-Net & 0.44 & 0.51 & 0.14 & 0.19 & 0.08 & 0.13 & 0.14 & 0.19 & 0.06 & 0.10 & 0.04 & 0.06 & 0.10 & 0.15 & 0.03 & 0.05 & 0.13 & 0.17 \\
MVSS-Net & 0.40 & 0.48 & 0.23 & 0.31 & 0.48 & 0.61 & 0.24 & 0.29 & 0.23 & 0.34 & 0.12 & 0.17 & 0.10 & 0.09 & 0.18 & 0.24 & 0.24 & 0.32 \\
FakeShield & \underline{0.54} & \underline{0.60} & \textbf{0.50} & \textbf{0.57} & \underline{0.67} & \underline{0.75} & \underline{0.32} & \underline{0.37} & \underline{0.48} & \underline{0.52} & \underline{0.17} & 0.20 & 0.14 & 0.22 & 0.18 & 0.24 & \underline{0.39} & \underline{0.45} \\
FakeShield* & 0.48 & 0.56 & \underline{0.47} & \underline{0.55} & 0.65 & 0.74 & 0.22 & 0.26 & 0.40 & 0.44 & 0.09 & 0.12 & \underline{0.25} & \underline{0.24} & \underline{0.17} & \underline{0.22} & 0.34 & 0.39 \\
\midrule
SIDA-7B* & 0.43 & 0.49 & 0.41 & 0.49 & 0.57 & 0.67 & 0.25 & 0.30 & 0.31 & 0.36 & 0.10 & 0.15 & 0.31 & 0.42 & 0.51 & 0.60 & 0.36 & 0.43 \\
SIDA-13B* & 0.45 & 0.52 & 0.45 & 0.52 & 0.60 & 0.71 & 0.27 & 0.32 & 0.31 & 0.35 & 0.12 & 0.17 & 0.31 & 0.42 & 0.53 & 0.61 & 0.38 & 0.45 \\

\midrule
IFDL-VLM (Ours) & \textbf{0.60} & \textbf{0.70} & 0.46 & \underline{0.55} & \textbf{0.77} & \textbf{0.88} & \textbf{0.33} & \textbf{0.43} & \textbf{0.55} & \textbf{0.67} & \textbf{0.19} & \textbf{0.29} & \textbf{0.39} & \textbf{0.47} & \textbf{0.47} & \textbf{0.61} & \textbf{0.47} & \textbf{0.58} \\
\bottomrule
\end{tabular}
}
\vspace{-0.1in}
\end{table*}

\noindent
\textbf{Localization Results.} 
Table~\ref{tab:localization_results} reports pixel-level forgery localization results on SID-Set. Our method achieves an AUC of 0.99, an F1 score of 0.87, and an IoU of 0.65, substantially outperforming all prior methods. Notably, we obtain a \textbf{+21\%} absolute improvement in IoU over the strongest baseline (from 0.44 to 0.65).

Overall, our model significantly surpasses prior methods in detection and localization, demonstrating the effectiveness of the proposed decoupled optimization between detection\&localization and language explanation generation.

\noindent
\textbf{Qualitative Comparison.} 
Figure~\ref{fig:visual_comparison} provides visual examples, showing that our method generates cleaner and more precise localization masks than SIDA, aligning closely with ground-truth tampered regions. Additionally, the generated language explanations are more faithful and detailed, explicitly describing the tampered regions and their semantic context, while SIDA often produces vague or partially incorrect rationales. These improvements highlight the advantages of decoupling detection\&localization from explanation and leveraging localization masks as auxiliary signals for interpretability. Please refer to Appendix~\ref{sec:qualitative_examples} for more visual examples.

\subsubsection{Evaluation on Cross-Dataset Generalization}
\label{subsec:fakeshield_setting}
We also evaluate the cross-dataset generalization performance to show the effectiveness of our method. Specifically, following previous work, we train our model on CASIAv2, FantasticReality, DeepFake, and AIGC Editing data. Then the evaluation is performed on 8 datasets, including CASIA1+, IMD2020, Columbia, NIST, DSO, Korus, DeepFake, and AIGC Editing. For fair comparisons, we keep consistent settings with FakeShield.

\noindent{\bf Training Setup.}
For fair comparisons, we follow FakeShield to train and evaluate our models on MMTD-Set with the same settings. 
Specifically, we optimize the model using AdamW optimizer with an initial learning rate of $1e-5$, $\beta=(0.9,0.95)$, and no weight decay. 
The learning rate linearly increases from 0 to 1e-5
 over the first 10\% of total training steps, followed by a linear decay to 0 over 12 epochs. We use a batch size of 32 with a gradient accumulation factor of 10 (yielding an effective global batch size of 320 across 4 GPUs). Mixed-precision training (FP16/BF16) is employed for efficiency, and gradient clipping is applied with a threshold of 1.0.

\noindent{\bf Comparison Results.} The experimental results are summarized in Table~\ref{FS_location_result_table},where we additionally retrain SIDA under the same setting for a fair comparison.  We obtain the best performance on 7 out of 8 datasets. Specifically, our model achieves an average IoU of 0.47 and an average F1 score of 0.58, outperforming the previous best method by $\textbf{13}\%$ IoU and $\textbf{19}\%$ F1 score, respectively. It again confirms the effectiveness of our method.

\begin{table}[t]
\centering
\caption{Quantitative comparison between SIDA-13B-description and our model on six evaluation dimensions with GPT-5. The overall score is computed as the average of the Mask score and the Text score.}
\vspace{-0.1in}
\resizebox{\linewidth}{!}{
\begin{tabular}{lcc}
\toprule
Section & SIDA-13B-description & IFDL-VLM (Ours) \\
\midrule
Mask & 1.22 $\pm$ 1.10 & \textbf{2.28 $\pm$ 1.68} \\
Type & 2.96 $\pm$ 2.44 & \textbf{3.70 $\pm$ 2.21} \\
Areas & 1.67 $\pm$ 1.40 & \textbf{2.45 $\pm$ 1.69} \\
Tampered Content & 1.14 $\pm$ 1.44 & \textbf{1.98 $\pm$ 1.80} \\
Visual Inconsistencies & 2.40 $\pm$ 1.04 & \textbf{2.91 $\pm$ 1.22} \\
Summary & 2.48 $\pm$ 2.15 & \textbf{3.31 $\pm$ 2.11} \\
\midrule
$\text{Overall}_{\text{text}}$ & 1.67 $\pm$ 1.16 & \textbf{2.44 $\pm$ 1.46} \\
Overall & 1.44 $\pm$ 1.04 & \textbf{2.36 $\pm$ 1.48} \\
\bottomrule
\end{tabular}}
\label{tab:gpt_eval}
\vspace{-0.25in}
\end{table}

\subsection{Evaluation on Language Interpretability}
\label{sec:comparison_mllm}

Beyond detection and localization, our IFDL-VLM framework is designed to provide faithful explanations for predicted tampered regions. To assess interpretability, we conduct both quantitative evaluations (user study, GPT-5 assessment and CSS) and qualitative case studies (in 
Appendix~\ref{sec:qualitative_examples} on SID-Set-description validation data.

\begin{figure}[t!]
\centering
\includegraphics[width=\linewidth]{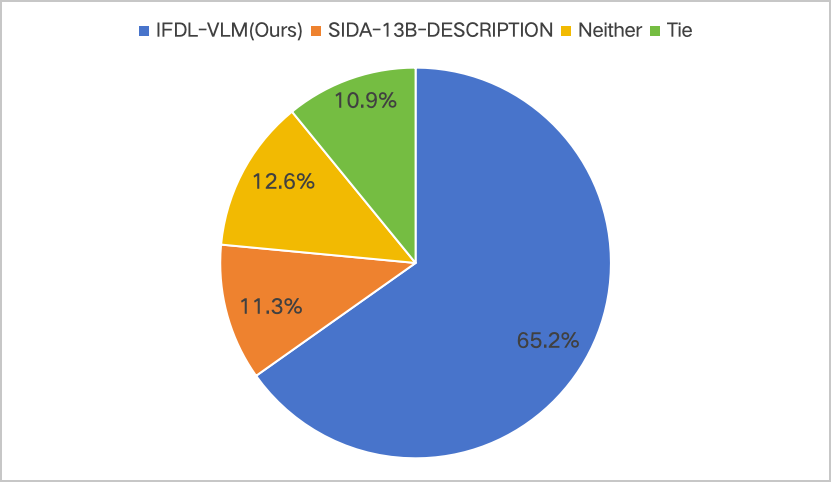}
\caption{Overall preference distribution in human evaluation.}
\label{fig:votes}
\vspace{-0.1in}
\end{figure}

\noindent{\bf User Study.}
\label{subsec:human_eval}
To quantitatively compare the interpretability of our model with the SIDA-13B model, we conduct a user study. Since the validity of language explanations depends on their correspondence to the predicted localization, we jointly evaluate the quality of both textual descriptions and predicted masks. Specifically, 50 raters participate in the evaluation using 40 images randomly sampled from the SID-Set validation set.
For each image, raters are shown the two systems’ binary masks and textual explanations in random order (to mitigate position bias) and asked to select \emph{one} of four options:
(i) \textbf{Model A better}: A’s mask and explanation are closer to the ground truth and overall more convincing; 
(ii) \textbf{Model B better}; 
(iii) \textbf{Neither}: both are far from the ground truth; 
(iv) \textbf{Tie}: both are similarly close to the ground truth.

We visualize the overall preference distribution via a pie chart (Figure~\ref{fig:votes}), which aggregates all judgments collected. The four slices represent the proportions of preferences for our model, SIDA, ``neither'', and ``tie''. 
Notably, our method is preferred in \textbf{65.2\%} of all comparisons, substantially outperforming SIDA-13B (11.3\%), while 12.6\% of cases are judged as ``neither'' and 10.9\% as ``tie''. 
This overwhelming preference confirms that our explanations are significantly more faithful and convincing to human evaluators.


\noindent{\bf GPT-5 Evaluation.}
\label{subsec:gpt_eval}
Following prior work FakeShield, we employ a multimodal large language model (MLLM) judge (GPT-5), which is provided with the tampered image, ground-truth (GT) mask, two predicted masks (ours vs.\ SIDA), their textual explanations, and the GT rationale. The judge is instructed to assign per-dimension scores in $[0,5]$ across six aspects: mask prediction (localization fidelity to GT), forgery type, tampered areas, tampered content, visual inconsistencies, and summary.

As summarized in Table~\ref{tab:gpt_eval}, our IFDL-VLM model consistently outperforms the SIDA-13B description baseline in all six aspects, demonstrating improved multimodal reasoning and coherence with the results of detection and localization. 
To mitigate the potential bias from generic or templated responses—particularly in high-level categories like \textit{type} and \textit{summary}, which may achieve spuriously high scores via surface-level phrasing—we assign lower weights to these dimensions. Specifically, guided by the principle that interpretable forgery analysis should prioritize concrete ``where'', ``what'', and ``why'' (i.e., spatial location, semantic content, and visual anomalies), we integrate the scores of forgery type, tampered areas, tampered content, visual inconsistencies, and summary with the weights $\{0.05, 0.35, 0.40, 0.15, 0.05\}$, yielding an overall language-only score of 2.44, surpassing SIDA-13B by approximately \textbf{46\%}. When combining both mask and textual scores, IFDL-VLM achieves an overall score of 2.36 versus 1.44 for SIDA-13B, further validating the advantage of IFDL-VLM regarding interpretability.



\noindent{\bf CSS Evaluation.}
Additionally, we assess textual faithfulness using cosine semantic similarity (CSS) with all-mpnet-base-v2. Similar to GPT-5 evaluation, we report CSS scores in 5 aspects covering tampered type, tampered area, tampered content, visual inconsistencies and summary. To emphasize ``where'', ``what'' and ``why'', the CSS(weight) score is calculated in a weighted sum manner using the weights $\{0.05,0.35,0.40,0.15,0.05\}$.

As shown in Table~\ref{tab:css_sections}, our model consistently performs much better in all five aspects, validating the effectiveness of our method in terms of language interpretability.  Specifically, attaching more importance on aspects of tampered area, tampered content, and visual inconsistencies, we achieve a CSS(weighted) score of 0.62, which largely surpasses the SIDA-13B by \textbf{8.8\%}.

\begin{table}[t] 
\centering 
\caption{Section-wise CSS scores.} 
\vspace{-0.1in}
\label{tab:css_sections} 
\setlength{\tabcolsep}{10pt} 
\renewcommand{\arraystretch}{1.12} 
\resizebox{1.0\linewidth}{!} 
{ 
\begin{tabular}{lcc} 
\toprule 
Section & SIDA-13B-description & Ours \\ 
\midrule 
Type & $0.83$ & $\mathbf{0.87}$ \\ Areas & $0.61$ & $\mathbf{0.67}$ \\
Tampered Content & $0.44$ & $\mathbf{0.49}$ \\ 
Visual Inconsistencies & $0.67$ & $\mathbf{0.70}$ \\ 
Summary & $0.80$ & $\mathbf{0.84}$ \\ 
CSS(weighted) & $0.57$ & $\mathbf{0.62}$ \\ 
\bottomrule 
\end{tabular} } 
\vspace{-0.25in} 
\end{table}

\section{Conclusion}
In conclusion, this work provides a comprehensive exploration of how Vision-Language Models (VLMs) can be effectively leveraged for the Image Forgery Detection and Localization (IFDL) task. We identify that vision-language alignment priors in existing VLMs are inherently biased toward semantic plausibility rather than authenticity, which limits their effectiveness in forgery analysis. To mitigate this, we decouple the optimization of detection\&localization from VLM-based language interpretability. With a trainable ViT backbone and a pretrained SAM, our approach achieves substantial improvements over previous methods in both detection and localization. Furthermore, we propose a region-aware visual feature enhancement strategy that utilizes localization masks to explicitly define forgery concepts, thereby facilitating VLM optimization and enhancing interpretability. Extensive experiments on 9 widely used benchmarks validate the generalization ability of the proposed IFDL-VLM framework across detection, localization, and interpretability.

{
    \small
    \bibliographystyle{ieeenat_fullname}
    \bibliography{main}
}

\clearpage

\appendix
\onecolumn
\begin{center}
	\Large \textbf{Rethinking VLMs for Image Forgery Detection and Localization}
	\Large \\ \textbf{Supplementary Material}
\end{center}
\vspace{20pt}

\section{Appendix / Supplemental Material}
\subsection{Ablation on $\alpha$ for Region-aware Visual Feature Enhancement}
\label{sec:alpha_ablation}

We conduct an ablation study to evaluate the effect of $\alpha$ in Eq.~(4): 
\[
T_{\text{vis}} = \alpha \text{CLIP}(x) + (1 - \alpha)  \text{CLIP}(x \odot M),
\]
where $\alpha$ controls the balance between the global semantics (from $\text{CLIP}(x)$) and region-specific forensic cues (from $\text{CLIP}(x \odot M)$, where $M$ is the mask). We evaluate $\alpha \in \{0, 0.3, 0.5, 0.7, 1.0\}$ and report performance on the SID-Set using the Cosine Semantic Similarity (CSS) score. 

The results show that $\alpha = 0.5$ achieves the best trade-off between global semantics and region-specific forensic cues, balancing the need for general image context with the local detail required for accurate forgery detection. Specifically, $\alpha = 0.5$ provides the highest CSS score, indicating a better alignment between the model’s predicted explanation and the ground-truth rationale.

Further analysis reveals that values of $\alpha$ closer to 0 (favoring global semantics) or closer to 1 (favoring localized features) lead to suboptimal performance, as they either lose important contextual understanding or miss crucial localized details. The results are summarized in Table \ref{tab:alpha_ablation_css}, which shows how varying $\alpha$ influences the model's overall performance.

\begin{table}[h]
\centering
\caption{Ablation study on $\alpha$ for region-aware visual
feature enhancement technique with CSS scores.}
\label{tab:alpha_ablation_css}
\setlength{\tabcolsep}{10pt}
\renewcommand{\arraystretch}{1.12}
\begin{tabular}{lccccc}
\toprule
\textbf{$\alpha$ Value} & \textbf{Type} & \textbf{Areas} & \textbf{Tampered Content} & \textbf{Visual Inconsistencies} & \textbf{Summary} \\
\midrule
0.0  & 0.79 & 0.64 & 0.54 & 0.59 & 0.77 \\
0.3  & 0.83 & 0.66 & 0.58 & 0.65 & 0.80  \\
0.5  & \textbf{0.87} & 0.67 & \textbf{0.60} & \textbf{0.70} & \textbf{0.84} \\
0.7  & 0.80 & \textbf{0.68} & 0.55 & 0.62 & 0.79 \\
1.0  & 0.75 & 0.65 & 0.50 & 0.60 & 0.77 \\
\bottomrule
\end{tabular}
\end{table}

\subsection{Effect of Unfreezing CLIP}
\label{sec:more exp about sida}
While the vision-language alignment in pre-trained VLMs aids in generating language explanations, these models inherently lack forgery-specific priors, such as sensitivity to inconsistent low-level cues. Furthermore, CLIP's shared image–text feature space is established through large-scale pre-training on up to 400M image–text pairs, acting as a crucial bridge for downstream reasoning tasks such as VQA. Directly fine-tuning the CLIP visual encoder on a relatively small image forgery dataset with localized segmentation losses disrupts this delicate cross-modal alignment. Consequently, this degradation severely impairs the model’s ability to generate language explanations. As shown in Table \ref{tab:sida_unfrozen_clip}, unfreezing CLIP in the SIDA baseline leads to noticeable drops in both explanation quality (CSS and ROUGE-L) and localization performance (IoU), further validating the necessity of our decoupled architecture.
\begin{table}[h]
\centering
\renewcommand{\arraystretch}{1.3} 
\setlength{\tabcolsep}{10pt}      

\caption{Comparison with SIDA (unfrozen CLIP).}
\label{tab:sida_unfrozen_clip}
\begin{tabular}{lccccc} 
\toprule
Method & AUC & F1 & IoU & CSS & ROUGE-L \\
\midrule
SIDA-7B                & 0.87 & 0.74 & 0.44 & 0.66 & 0.38 \\
SIDA-13B               & 0.94 & 0.72 & 0.54 & 0.80 & 0.41 \\
\midrule
SIDA-7B-Unfrozen-CLIP  & 0.94 & 0.63 & 0.40 & 0.65 & 0.39 \\
SIDA-13B-Unfrozen-CLIP & 0.95 & 0.72 & 0.50 & 0.73 & 0.40 \\
\midrule
\textbf{IFDL-VLM (Ours)} & \textbf{0.99} & \textbf{0.87} & \textbf{0.65} & \textbf{0.84} & \textbf{0.43} \\
\bottomrule
\end{tabular}
\end{table}

\subsection{Attention Module Details}
\label{sec:attention details}
\begin{figure}[h]
    \centering
    \includegraphics[width=0.9\linewidth]{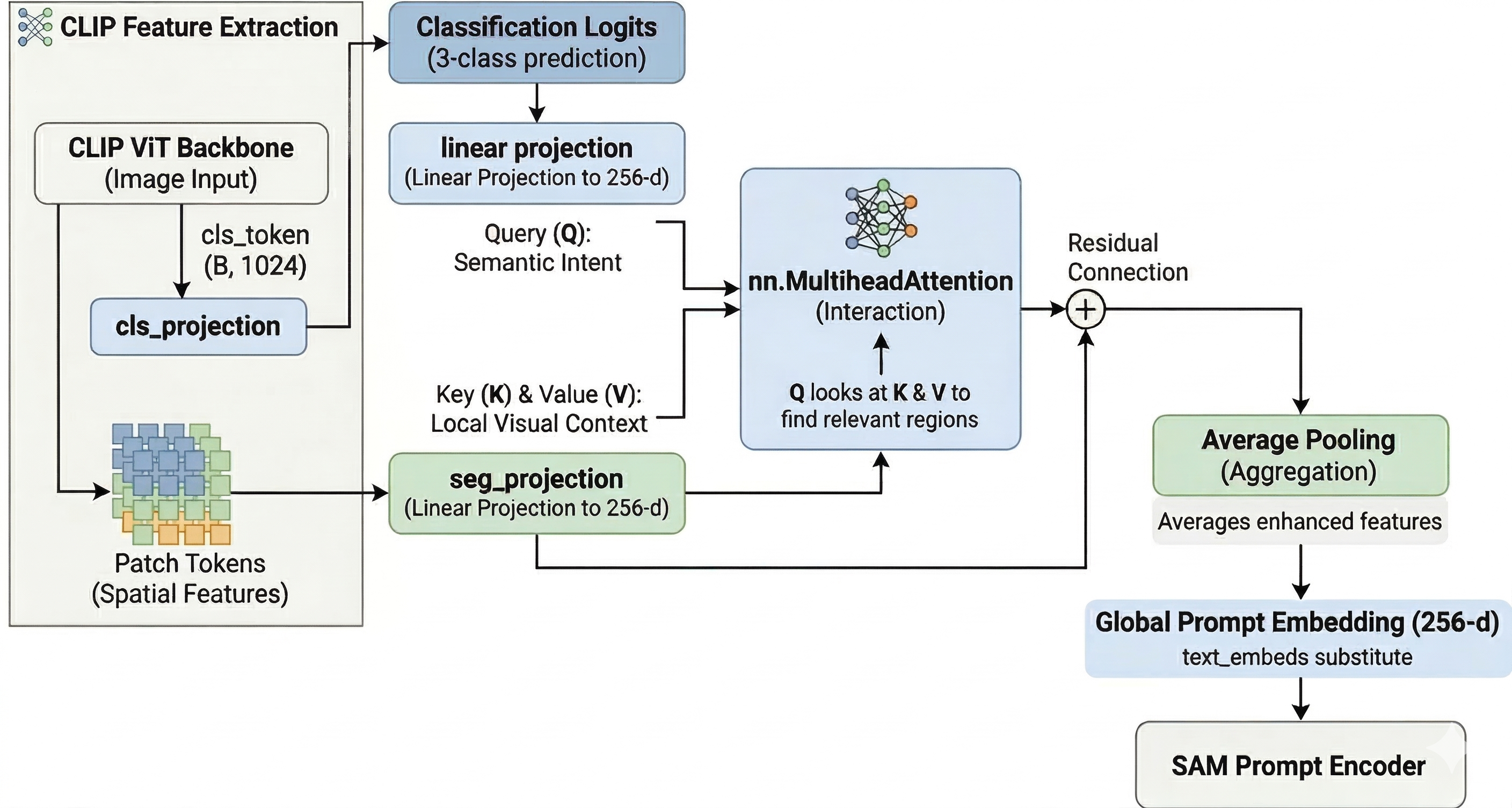}
    \caption{Attention details in the IFDL-VLM framework.}
    \label{fig:attention}
\end{figure}

As illustrated in Figure \ref{fig:attention}, the Attention Module functions as a Cross-Modal Feature Enhancement mechanism, effectively bridging global classification semantics with local visual features. The process begins with the CLIP ViT backbone extracting a global class token and spatial patch tokens from the input image. The class token is processed to generate classification logits, which represent the model's high-level semantic intent (e.g., the likelihood of tampering). To translate this semantic intent into spatial guidance for segmentation, the classification logits and the patch tokens are linearly projected into a shared 256-dimensional space. The projected logits serve as the Query ($Q$), while the projected patch tokens serve as the Key ($K$) and Value ($V$). Through a Multi-Head Attention layer, the semantic query interacts with the local visual context, dynamically highlighting regions relevant to the predicted class. The attention output is then added to the projected spatial features via a residual connection. Finally, an average pooling operation aggregates these enhanced features into a single 256-dimensional Global Prompt Embedding. This embedding acts as a substitute for text embeddings and is fed into the SAM Prompt Encoder to guide the mask decoder to focus specifically on the manipulated areas.

\subsection{Efficiency comparison}
\label{sec:Efficiency}
We report inference-time efficiency metrics (parameter count, FLOPs, and peak memory usage) and compare with SIDA and FakeShield under the same setting.
\begin{table}[h]
\centering
\renewcommand{\arraystretch}{1.3} 
\setlength{\tabcolsep}{12pt} 

\caption{Efficiency comparison.}
\label{tab:efficiency}
\begin{tabular}{lccc}
\toprule
Metric & FakeShield & IFDL-VLM & SIDA \\
\midrule
Params (B)         & 22.0 & 14.3 & 14.0 \\
FLOPs (T)          & 6.3  & 6.2  & 6.1 \\
Peak Mem. (GB)     & 27.1 & 27.5 & 29.5 \\
\bottomrule
\end{tabular}
\end{table}

\subsection{\bf Standard Text Similarity Metrics.} In addition to the model-based semantic similarity evaluations (GPT-5 and CSS), we also assess the generated explanations using standard natural language generation metrics, including BLEU-1, ROUGE-L, METEOR, and CIDEr. These metrics emphasize lexical overlap and n-gram matching. As shown in Table~\ref{tab:standard_metrics}, our IFDL-VLM consistently outperforms the SIDA-13B baseline across all traditional text similarity metrics. This further demonstrates that our framework not only aligns better with human semantics but also generates text that is lexically more faithful to the ground-truth annotations.

\begin{table}[h]
  \centering
  \small
  \setlength{\tabcolsep}{4pt}
  \begin{tabular}{l c c c c}
    \toprule
    Method & BLEU-1 & ROUGE-L & METEOR & CIDEr \\
    \midrule
    SIDA-13B & 0.548 & 0.406 & 0.272 & 0.016 \\
    IFDL-VLM (Ours) & \textbf{0.580} & \textbf{0.434} & \textbf{0.290} & \textbf{0.030} \\
    \bottomrule
  \end{tabular}
  \vspace{-2mm}
  \caption{Quantitative comparison on standard text similarity metrics.}
  \label{tab:standard_metrics}
\end{table}

\subsection{Automated GPT-5 Evaluation Protocol for IFDL Explanation.}
\label{sec:gpt5_eval}

To assess the quality of generated language explanations, we employ an automated evaluation protocol using the multimodal large language model (GPT-5). The input to GPT-5 consists of:  
(i) the tampered image,  
(ii) the ground-truth (GT) localization mask,  
(iii) two predicted masks from Model A and Model B, and  
(iv) their corresponding textual explanations, alongside a human-written reference rationale (GT text).

To ensure the reproducibility of our automated evaluation, the GPT-5 model is applied across all methods with fixed decoding parameters: temperature = 0.7, top-p = 0.95, and maximum output tokens = 8000.

GPT-5 evaluates both models on six criteria, using a 0–5 scale:  
\textbf{mask} (localization fidelity), \textbf{type} (forgery type), \textbf{areas} (tampered regions), \textbf{tampered} (tampered objects/parts), \textbf{visual} (consistency of visual evidence), and \textbf{summary} (overall summary accuracy). The final score is computed by aggregating the five textual dimensions with weights {0.05, 0.35, 0.40, 0.15, 0.05} to obtain a weighted text score, which is then fused with the mask score as:  
\[
\text{Overall} = 0.5 \cdot score_{Mask} + 0.5 \cdot Overall_{text}.
\]  
All results reported in Table 4 are averaged over all of SIDA validation images, with the mean ± standard deviation.

\begin{figure}[h]
    \centering
    \includegraphics[width=0.95\linewidth]{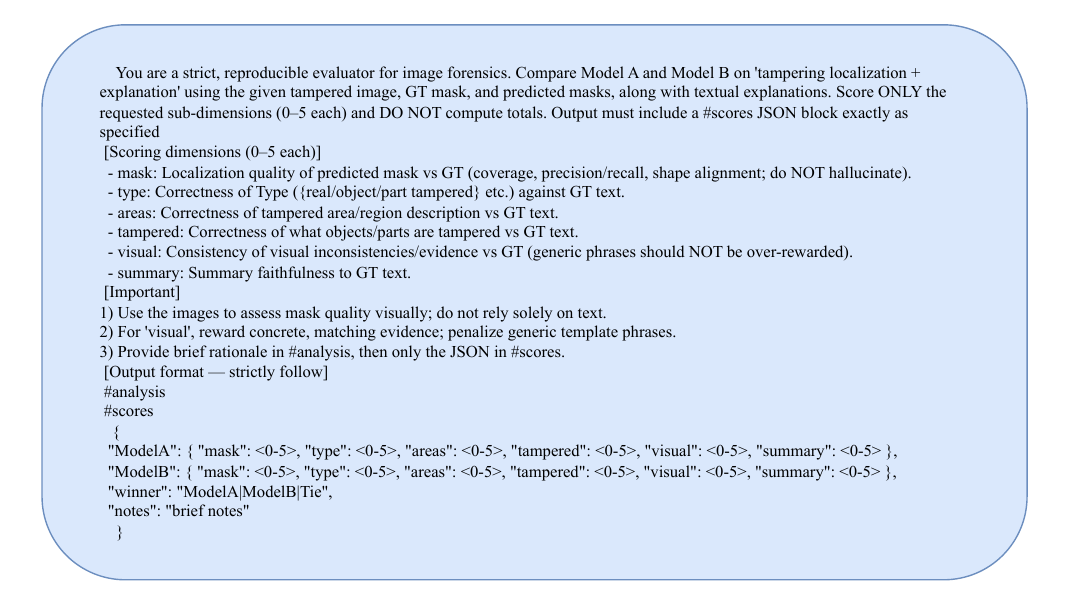}
    \caption{The full multimodal prompt used to instruct GPT-5 for pairwise evaluation of IFDL models. The prompt specifies scoring criteria, output format, and input modalities (image, masks, texts).}
    \label{fig:gpt5_prompt}
\end{figure}

The complete GPT-5 prompt used for this evaluation is shown in Figure~\ref{fig:gpt5_prompt}.

\subsection{Dataset Details}
\label{sec:dataset}
We used several publicly available datasets for both training and evaluation. Below are the details of the datasets, their composition, and how we used them.

\noindent\textbf{SID-Set}: The SID-Set consists of 300,000 images evenly split into three categories: Real, Full Synthetic, and Tampered images. We divide the dataset into training, validation, and test sets with a 7:1:2 ratio, ensuring balanced class distribution across all splits. The mask annotations are used as ground truth for evaluating the localization performance. This dataset was preprocessed using the setup detailed in the original paper.

\noindent\textbf{MMTD-Set}: The MMTD-Set is composed of 8 different evaluation benchmarks, including CASIA1+, IMD2020, Columbia, and others. It includes images of various forgery types, such as copy-move, splicing, and removal. All of these datasets provide masks, which we utilize for our evaluation of localization. 

\subsection{Discussion: Error Propagation and Hallucination}
\label{sec:error_analysis}

A natural question for a decoupled pipeline is whether errors in Stage-1 localization could propagate to Stage-2 and cause the LLM to generate hallucinated forgery explanations. In particular, if the localization module produces an incorrect mask, such as a false positive over an authentic region, one may wonder whether the LLM would simply follow the mask and describe non-existent manipulation artifacts.

\noindent\textbf{Secondary Verification by the LLM.} Our IFDL-VLM is designed to mitigate this risk. During Stage-2 training, the LLM is exposed to real, fully synthetic, and tampered images, together with their corresponding descriptive annotations. As a result, the model learns to verify the visual evidence within the prompted region, rather than assuming that the highlighted area must be manipulated solely because it is specified by the mask. Therefore, when presented with an erroneous mask covering a genuine region, the LLM can still produce authenticity-consistent explanations, such as noting that the lighting is natural and the boundaries are smooth without clear signs of manipulation. In this sense, the Stage-2 LLM serves as a secondary verification module that can reject incorrect localization cues and avoid hallucinating non-existent forensic traces.



\noindent\textbf{Robustness to Mask Imprecision.}
To further quantify the effect of imperfect Stage-1 localization on explanation generation, we conduct a boundary perturbation experiment. Specifically, we apply morphological dilation and erosion to the predicted masks to simulate imprecise localization. The results are reported in Table~\ref{tab:mask_perturb_text}.

Notably, the semantic similarity score (CSS) remains at 0.842 under boundary perturbation, which is very close to the performance obtained with ground-truth masks (0.853). The other text-generation metrics also exhibit only marginal changes. These results suggest that the Stage-2 LLM is not overly sensitive to small geometric deviations in the Stage-1 masks. Instead, it is able to focus on the salient visual inconsistencies within the indicated region while tolerating moderate boundary noise. Overall, these findings support that the proposed decoupled framework is robust to imperfect localization and effectively alleviates error propagation from Stage-1 to Stage-2.

\begin{table}[t]
\centering
\small
\setlength{\tabcolsep}{8pt}
\renewcommand{\arraystretch}{1.1}
\caption{Quantitative evaluation of explanation robustness under boundary-perturbed masks. ``Boundary-Pert.'' denotes applying morphological operations to the predicted masks.}
\label{tab:mask_perturb_text}
\begin{tabular}{l ccccc}
\toprule
Setting & BLEU-1 & ROUGE-L & METEOR & CIDEr & CSS \\
\midrule
PM (Clean)          & \textbf{0.579} & \textbf{0.434} & 0.290              & \textbf{0.030} & \underline{0.842} \\
PM (Boundary-Pert.) & \underline{0.574} & \underline{0.431} & \textbf{0.292} & 0.020          & 0.842 \\
GT                  & 0.565          & 0.428          & \underline{0.291} & \underline{0.026} & \textbf{0.853} \\
\bottomrule
\end{tabular}
\end{table}

\subsection{More Qualitative Comparison Examples}
\label{sec:qualitative_examples}
\begin{figure}[h]
    \centering
    \includegraphics[width=0.85\linewidth]{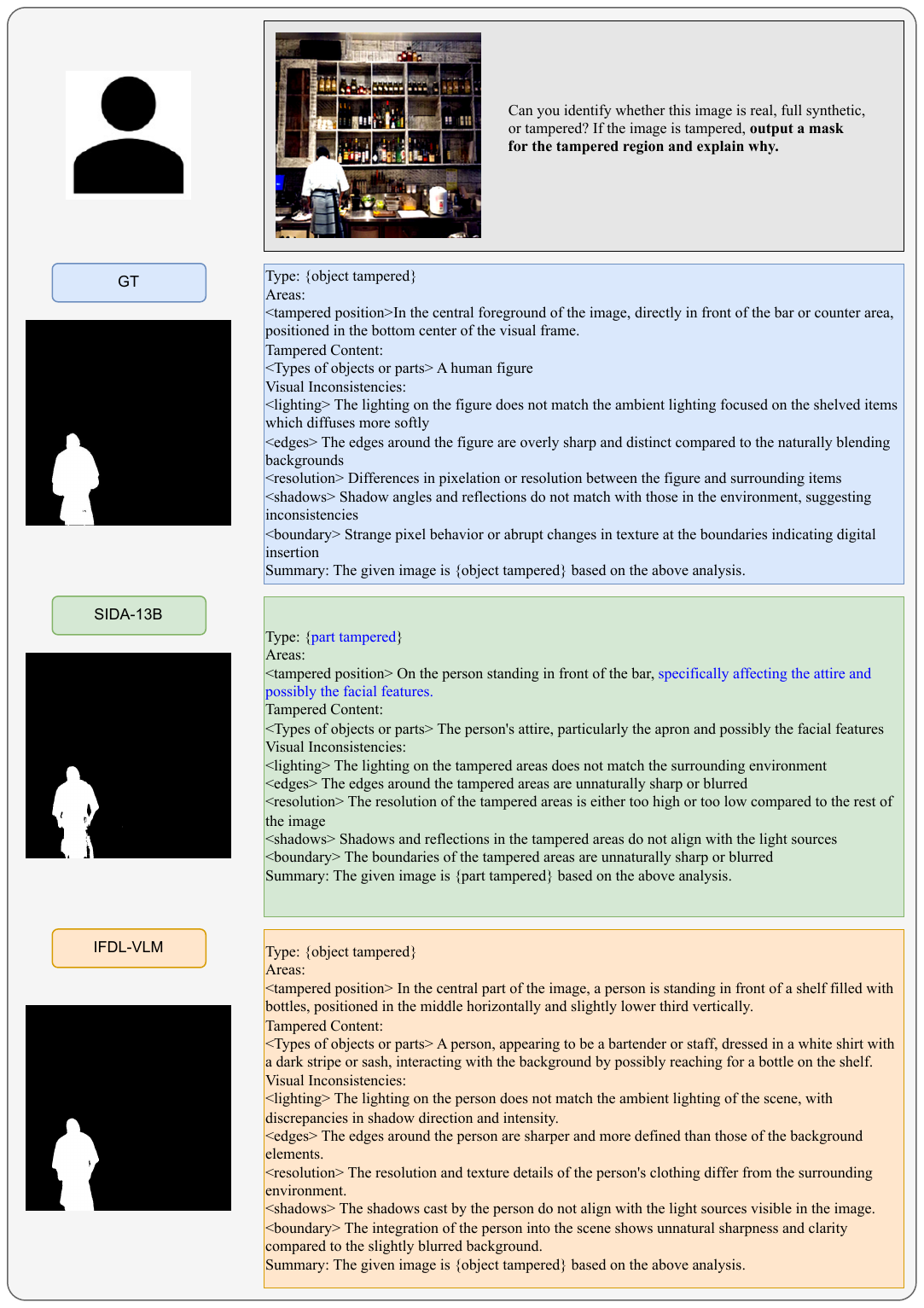}
    \caption{Qualitative comparison example. We highlight the accurate descriptions from GT in \textcolor{red}{red}, while the areas where the model fails to identify the modified regions correctly are marked in \textcolor{blue}{blue}. }
    \label{fig:example1}
\end{figure}

\begin{figure}[h]
    \centering
    \includegraphics[width=0.85\linewidth]{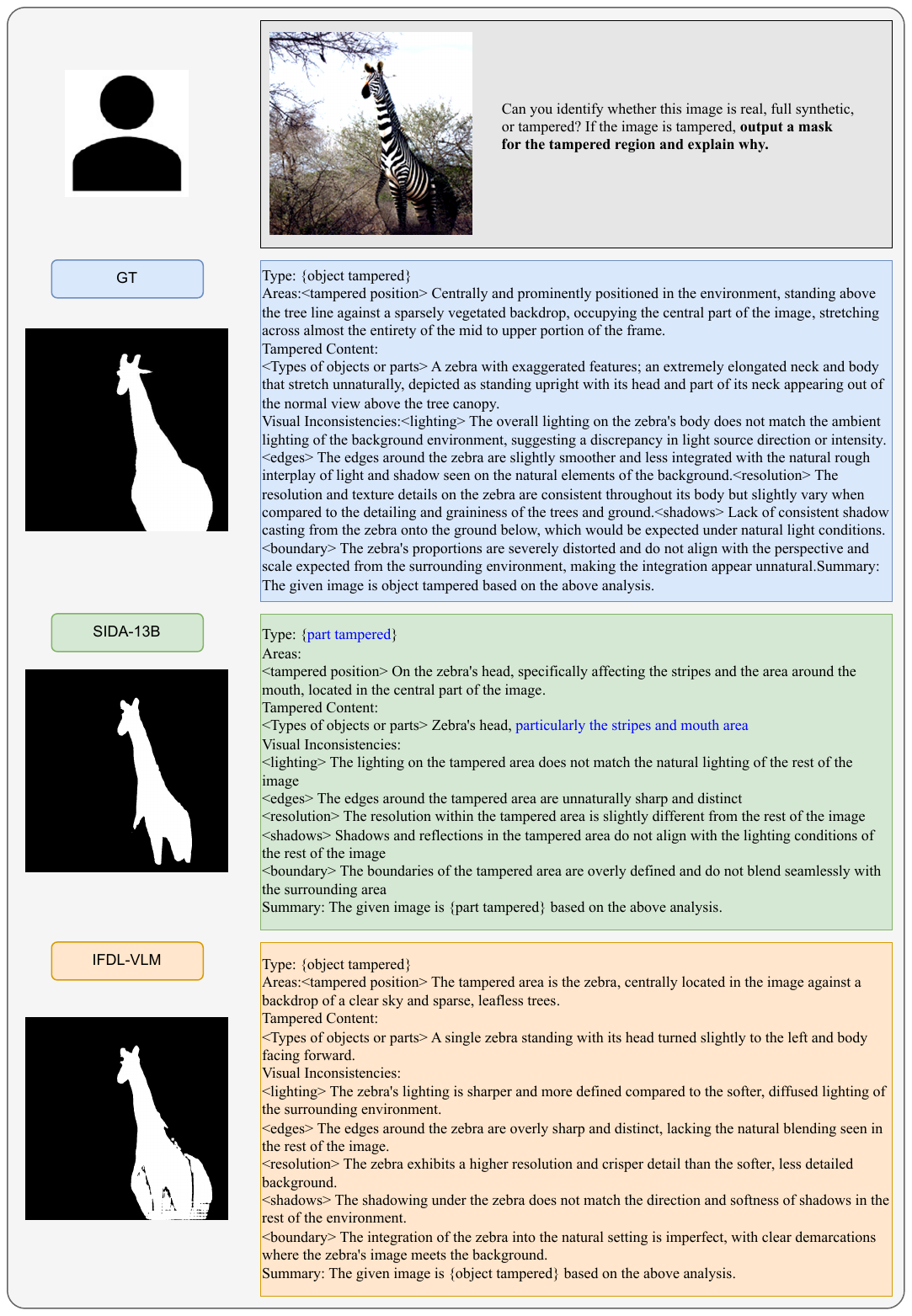}
    \caption{Qualitative comparison example. We highlight the accurate descriptions from GT in \textcolor{red}{red}, while the areas where the model fails to identify the modified regions correctly are marked in \textcolor{blue}{blue}. }
    \label{fig:example2}
\end{figure}

\begin{figure}[h]
    \centering
    \includegraphics[width=0.85\linewidth]{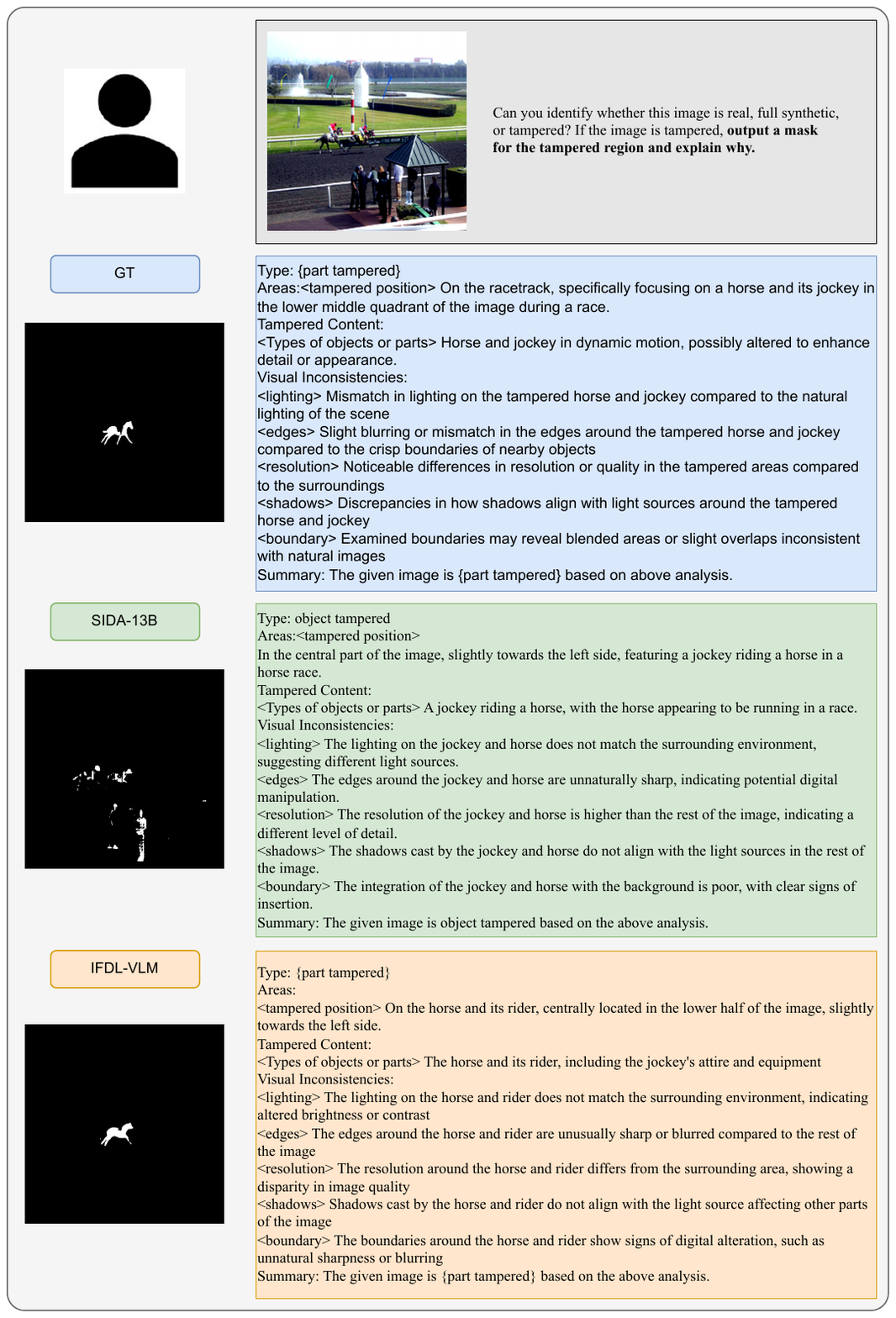}
    \caption{Qualitative comparison example. We highlight the accurate descriptions from GT in \textcolor{red}{red}, while the areas where the model fails to identify the modified regions correctly are marked in \textcolor{blue}{blue}. }
    \label{fig:example3}
\end{figure}

\begin{figure}[h]
    \centering
    \includegraphics[width=0.85\linewidth]{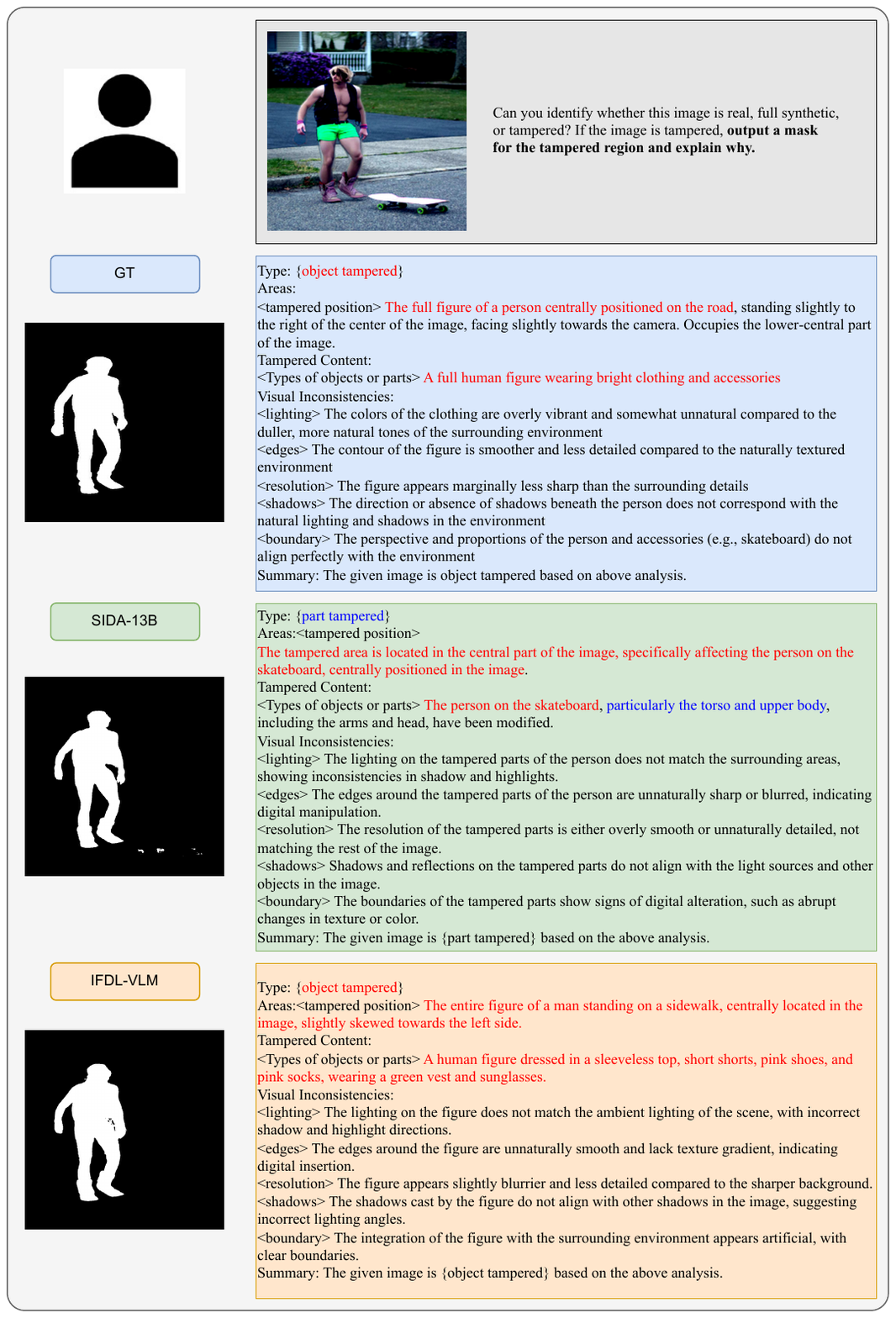}
    \caption{Qualitative comparison example. We highlight the accurate descriptions from GT in \textcolor{red}{red}, while the areas where the model fails to identify the modified regions correctly are marked in \textcolor{blue}{blue}. }
    \label{fig:example4}
\end{figure}

\begin{figure}[h]
    \centering
    \includegraphics[width=0.85\linewidth]{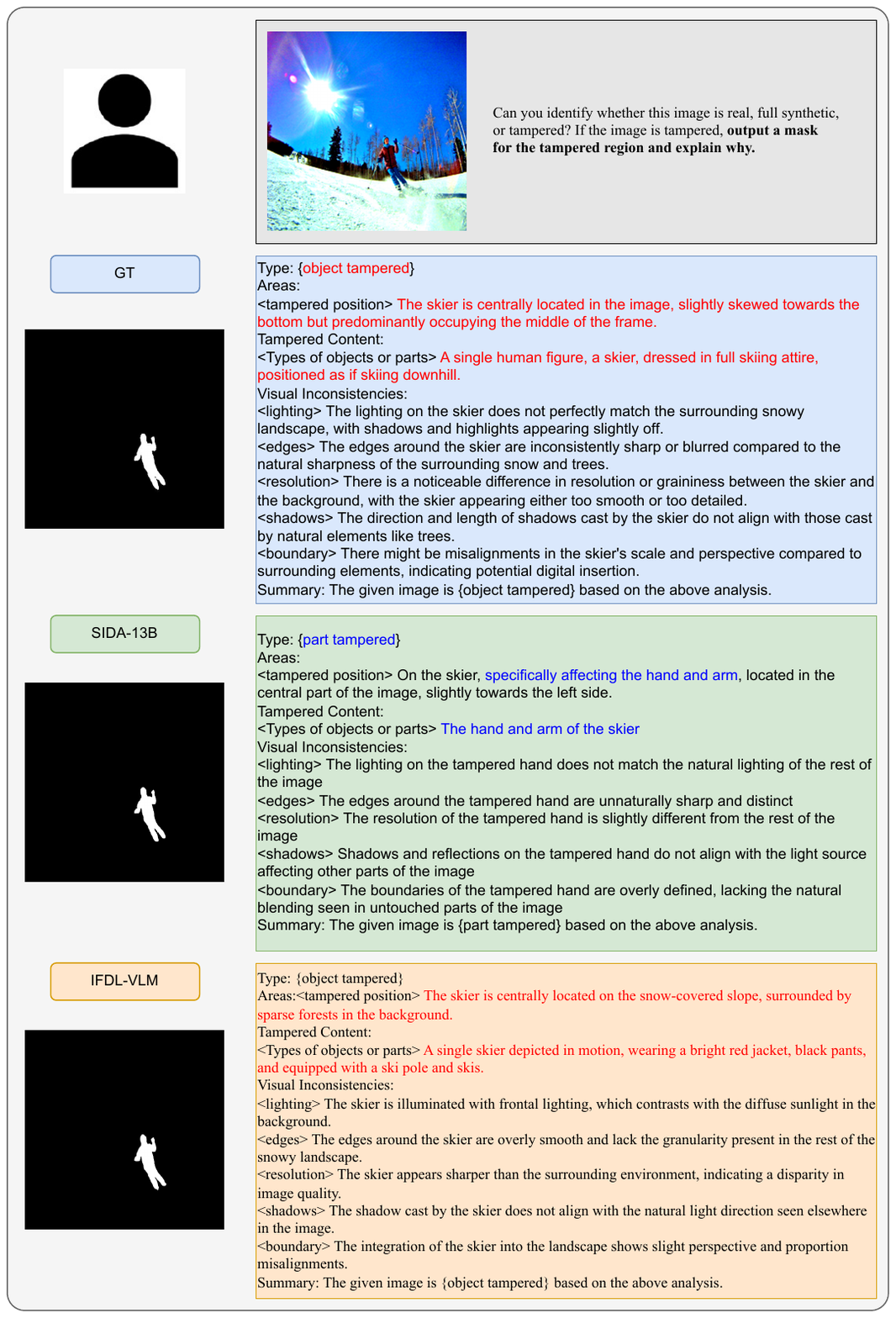}
    \caption{Qualitative comparison example. We highlight the accurate descriptions from GT in \textcolor{red}{red}, while the areas where the model fails to identify the modified regions correctly are marked in \textcolor{blue}{blue}. }
    \label{fig:example5}
\end{figure}

\begin{figure}[h]
    \centering
    \includegraphics[width=0.85\linewidth]{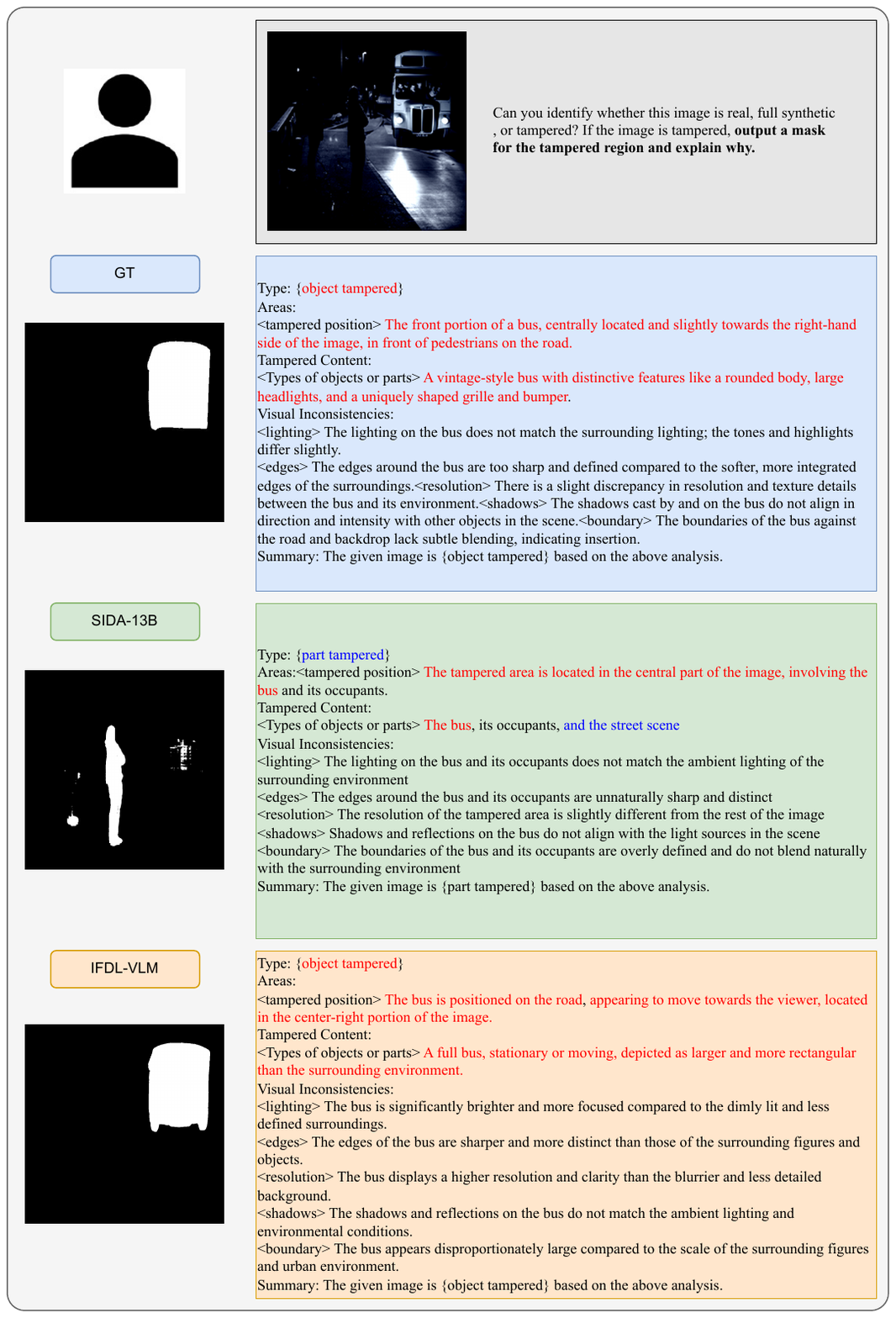}
    \caption{Qualitative comparison example. We highlight the accurate descriptions from GT in \textcolor{red}{red}, while the areas where the model fails to identify the modified regions correctly are marked in \textcolor{blue}{blue}. }
    \label{fig:example6}
\end{figure}

\begin{figure}[h]
    \centering
    \includegraphics[width=0.85\linewidth]{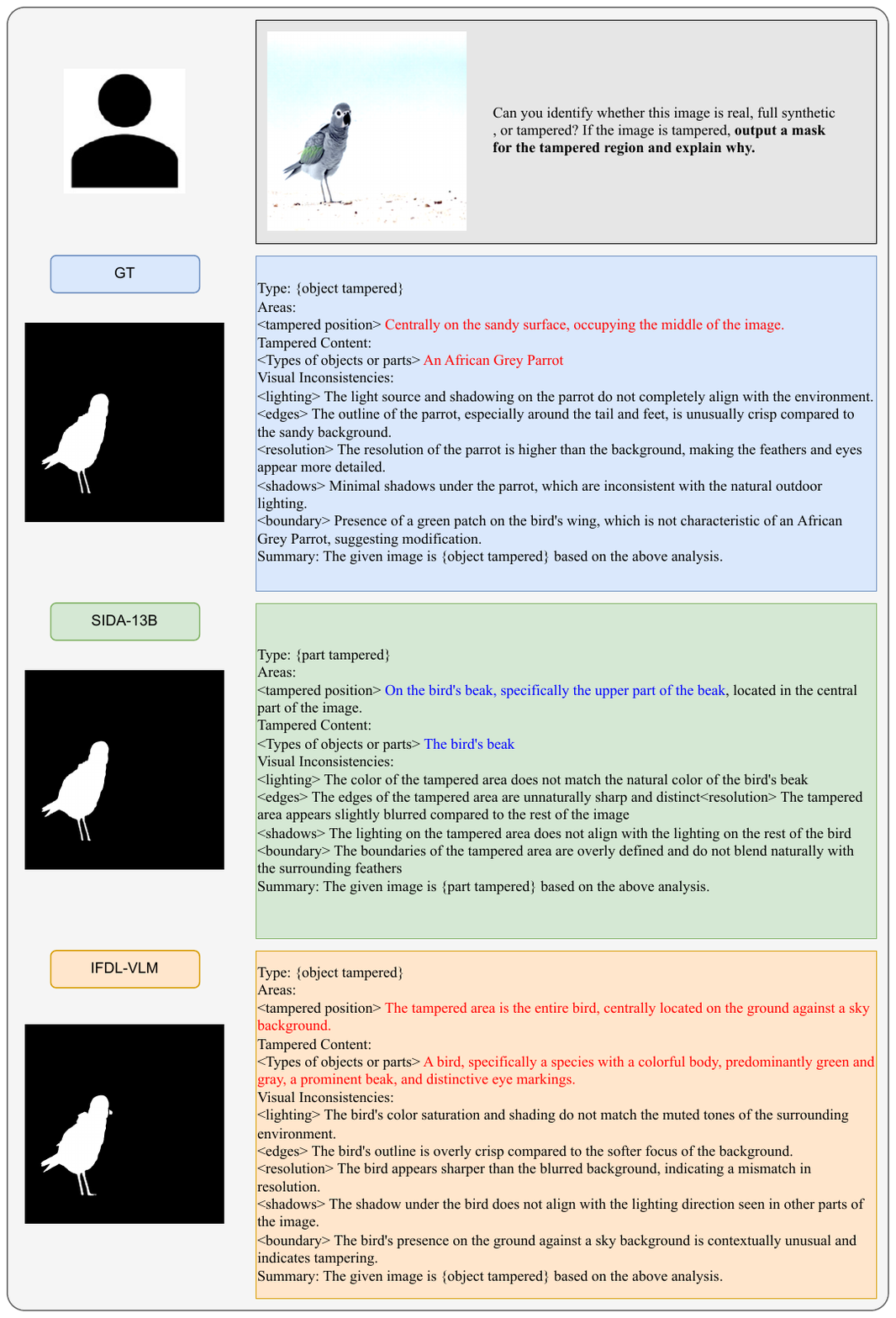}
    \caption{Qualitative comparison example. We highlight the accurate descriptions from GT in \textcolor{red}{red}, while the areas where the model fails to identify the modified regions correctly are marked in \textcolor{blue}{blue}. }
    \label{fig:example7}
\end{figure}

\begin{figure}[h]
    \centering
    \includegraphics[width=0.85\linewidth]{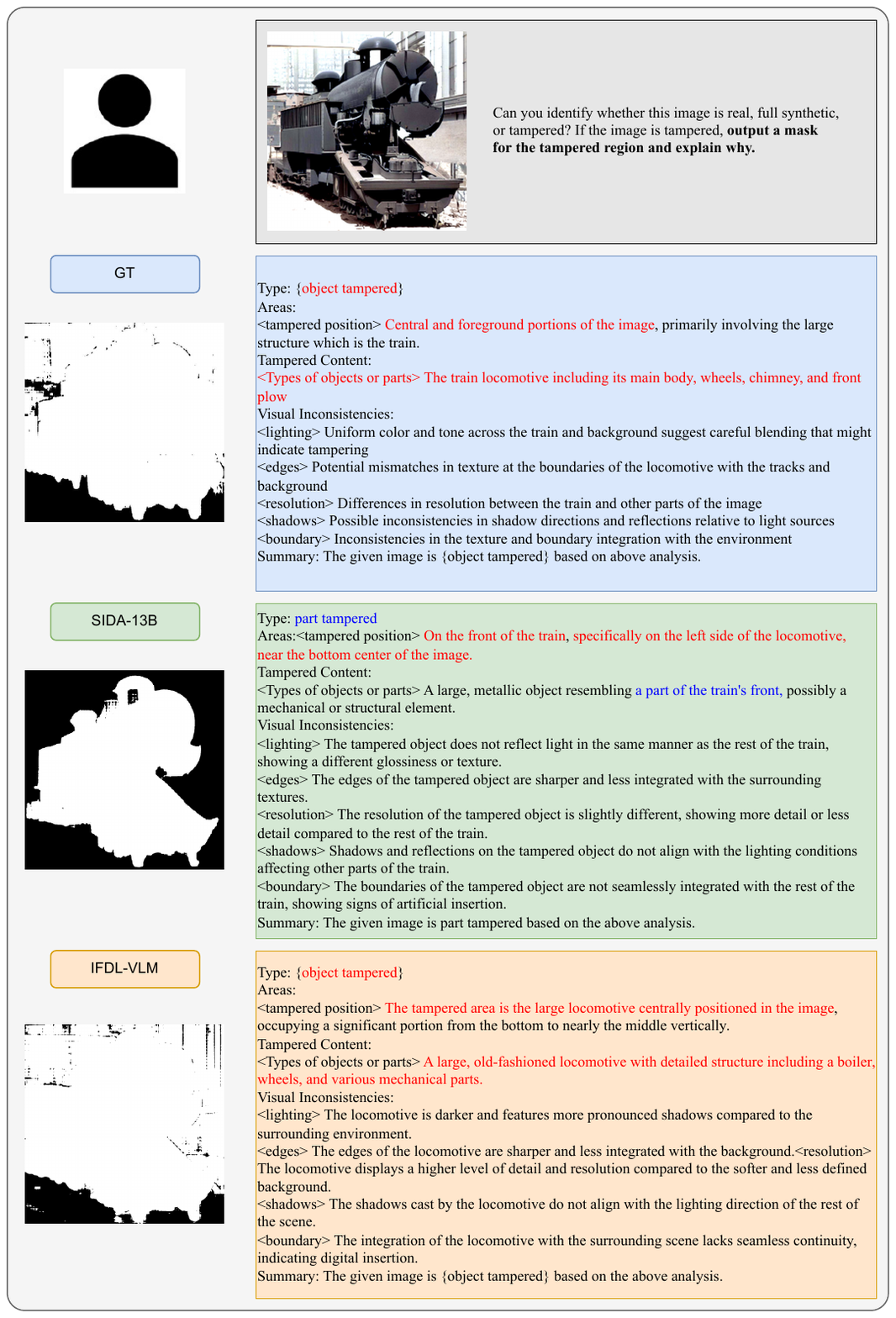}
    \caption{Qualitative comparison example. We highlight the accurate descriptions from GT in \textcolor{red}{red}, while the areas where the model fails to identify the modified regions correctly are marked in \textcolor{blue}{blue}. }
    \label{fig:example8}
\end{figure}

\begin{figure}[h]
    \centering
    \includegraphics[width=0.85\linewidth]{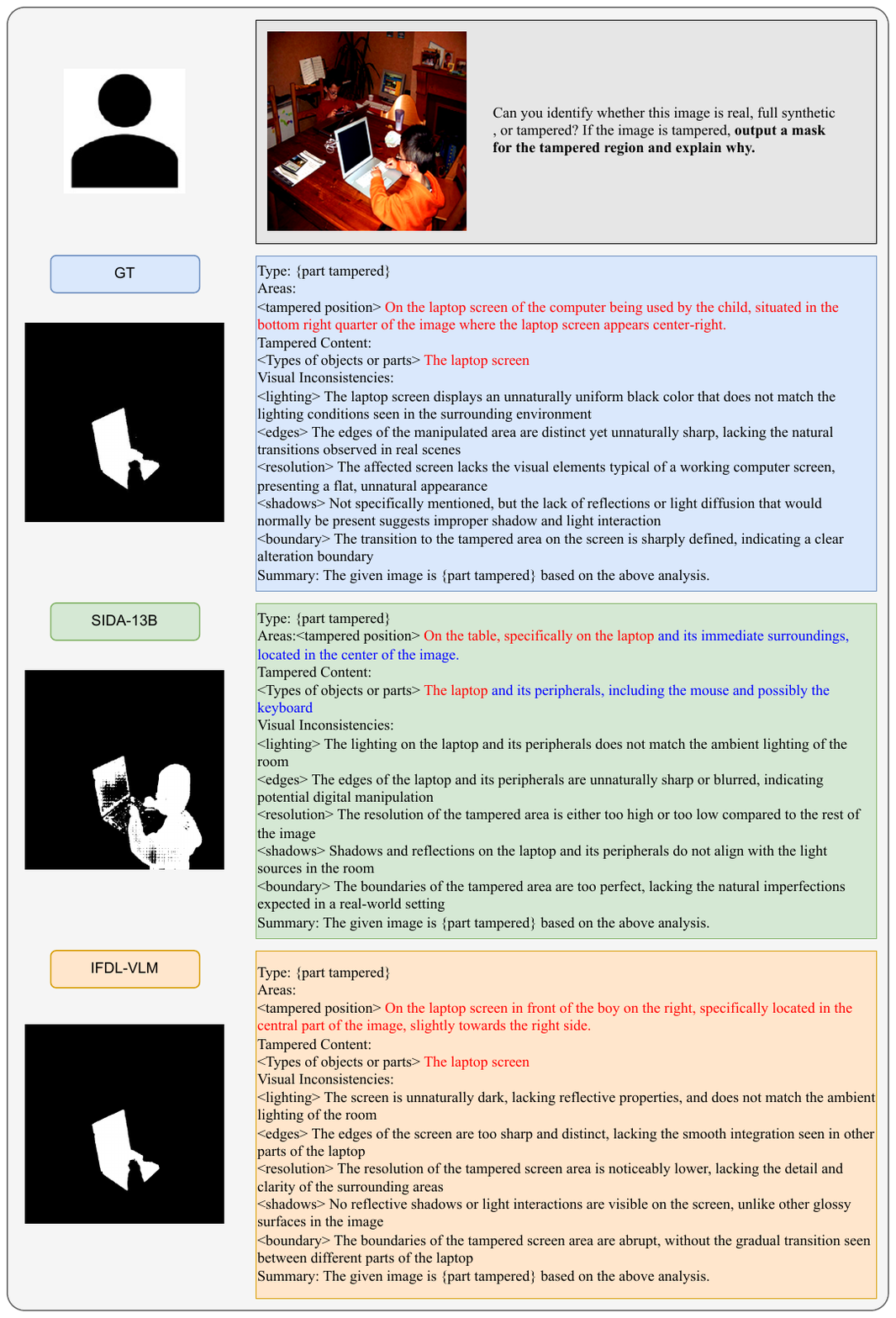}
    \caption{Qualitative comparison example. We highlight the accurate descriptions from GT in \textcolor{red}{red}, while the areas where the model fails to identify the modified regions correctly are marked in \textcolor{blue}{blue}. }
    \label{fig:example9}
\end{figure}

\begin{figure}[h]
    \centering
    \includegraphics[width=0.85\linewidth]{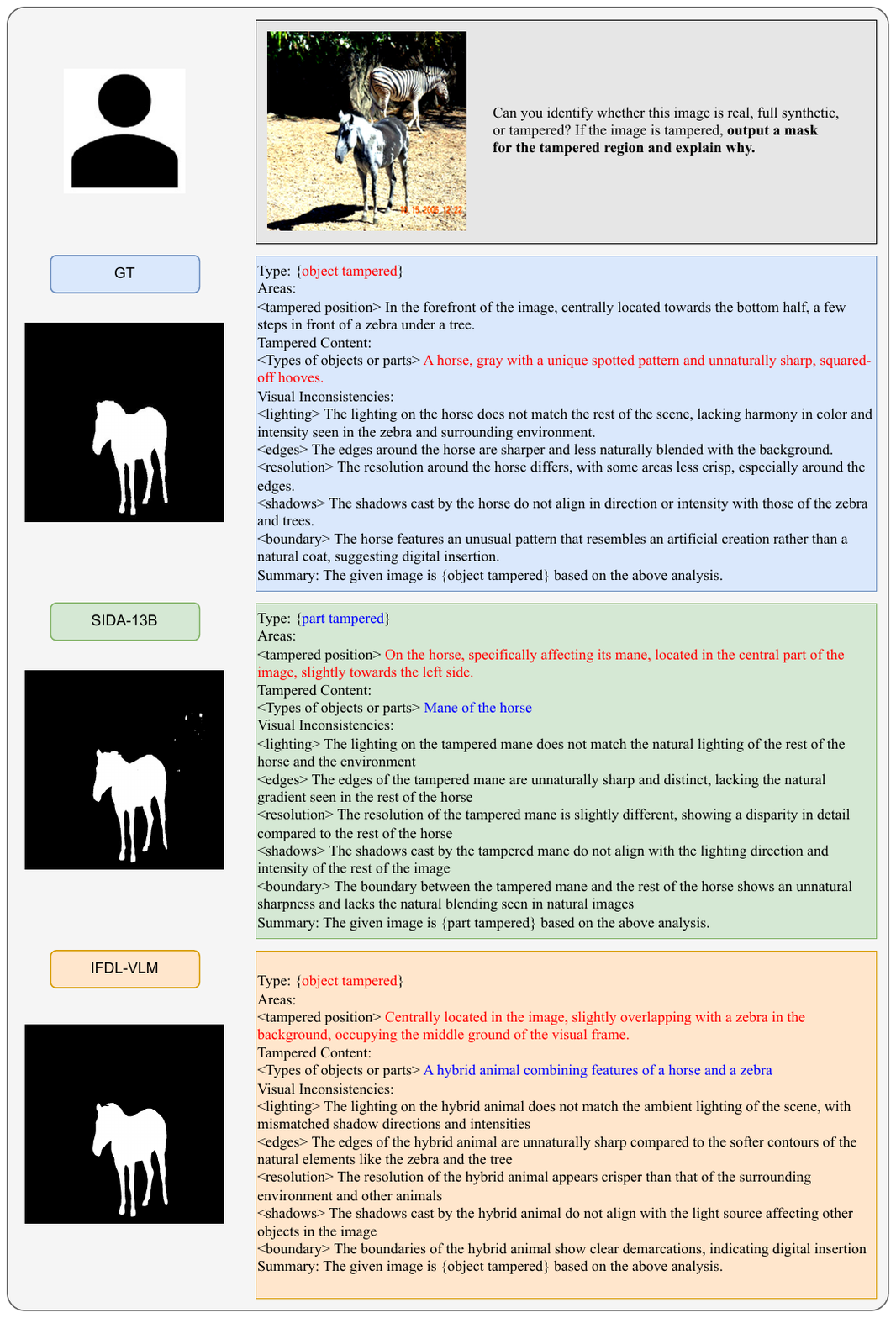}
    \caption{Qualitative comparison example. We highlight the accurate descriptions from GT in \textcolor{red}{red}, while the areas where the model fails to identify the modified regions correctly are marked in \textcolor{blue}{blue}. }
    \label{fig:example10}
\end{figure}

\begin{figure}[h]
    \centering
    \includegraphics[width=0.85\linewidth]{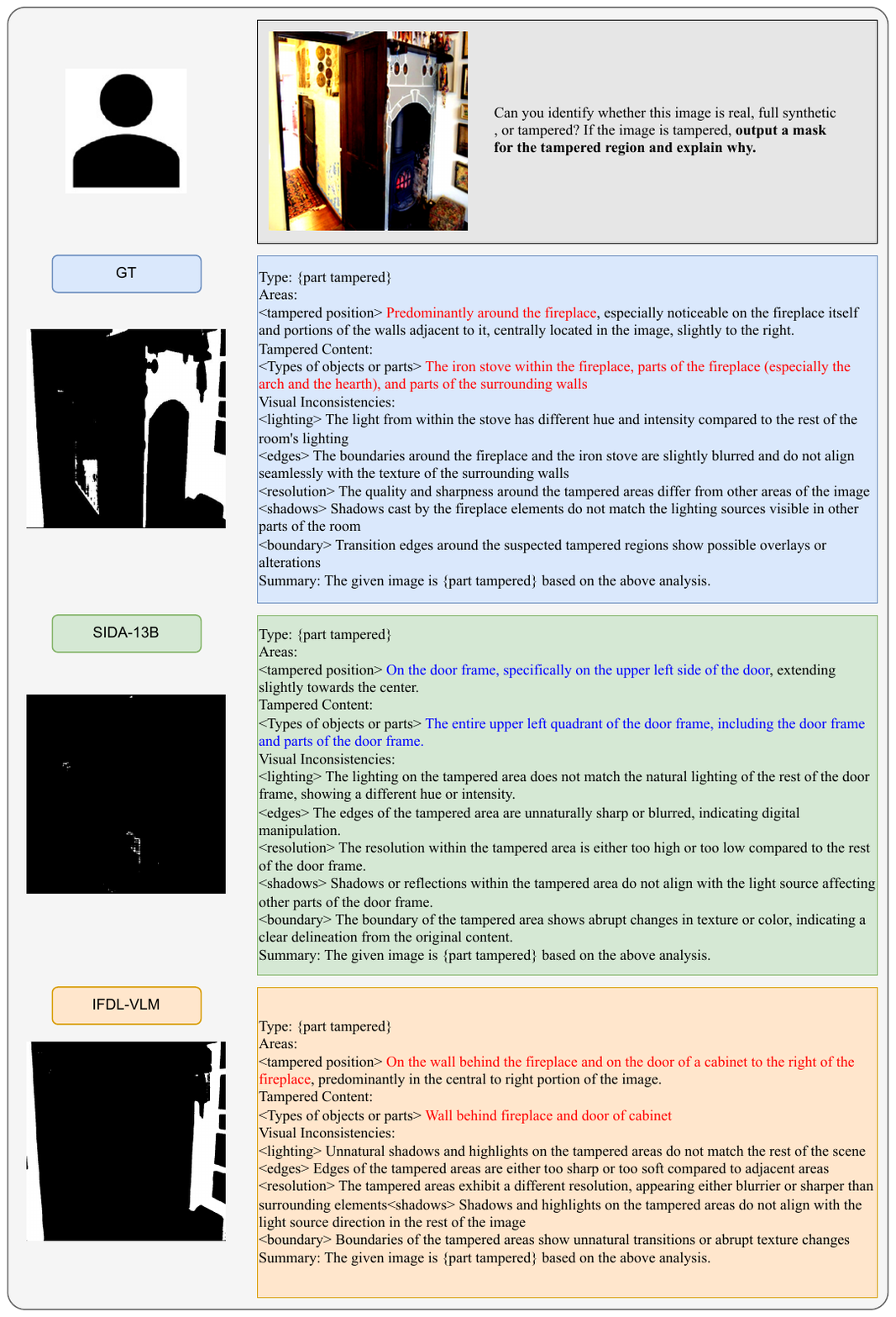}
    \caption{Qualitative comparison example. We highlight the accurate descriptions from GT in \textcolor{red}{red}, while the areas where the model fails to identify the modified regions correctly are marked in \textcolor{blue}{blue}. }
    \label{fig:example11}
\end{figure}

\begin{figure}[h]
    \centering
    \includegraphics[width=0.85\linewidth]{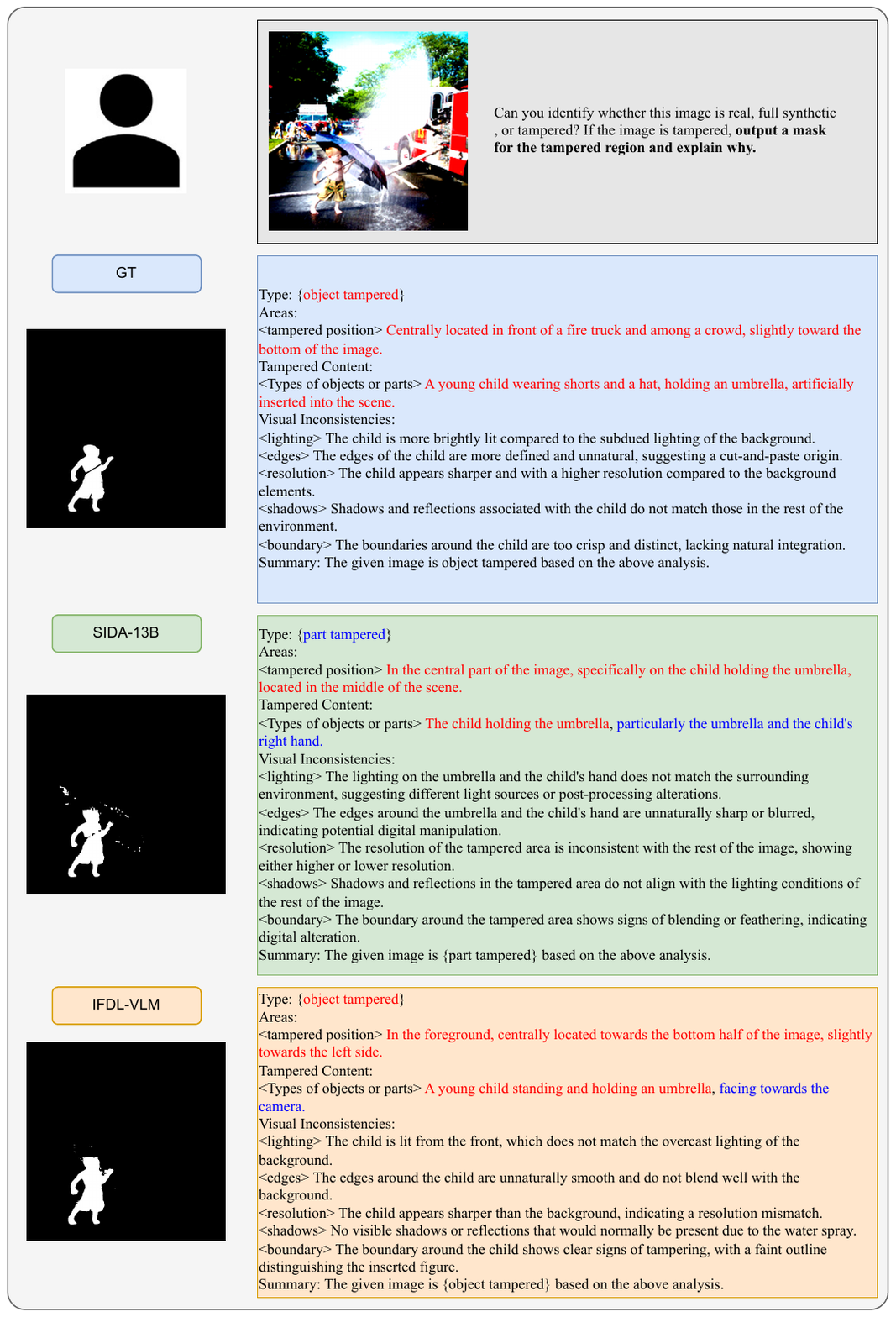}
    \caption{Qualitative comparison example. We highlight the accurate descriptions from GT in \textcolor{red}{red}, while the areas where the model fails to identify the modified regions correctly are marked in \textcolor{blue}{blue}. }
    \label{fig:example12}
\end{figure}

\begin{figure}[h]
    \centering
    \includegraphics[width=0.85\linewidth]{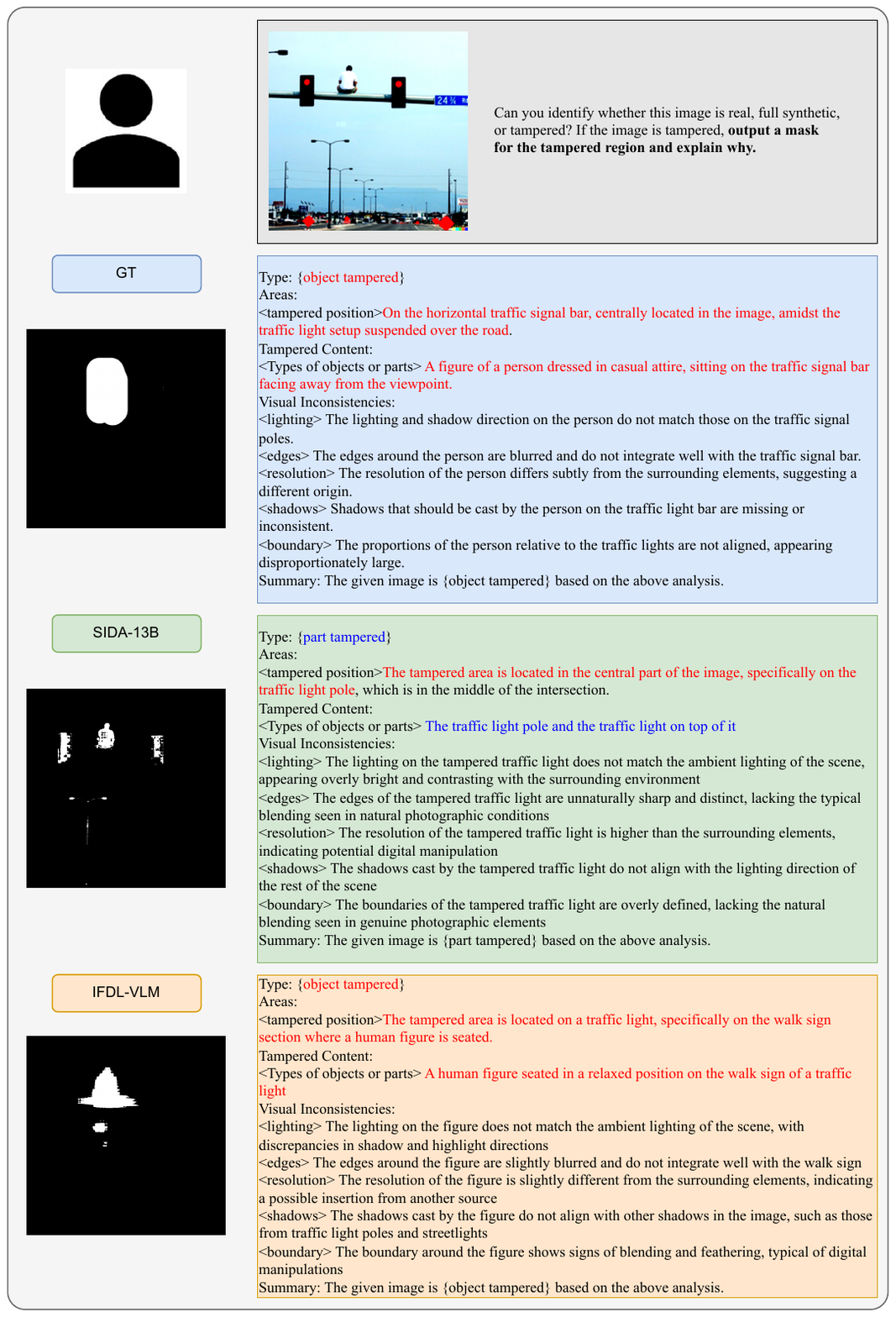}
    \caption{Qualitative comparison example. We highlight the accurate descriptions from GT in \textcolor{red}{red}, while the areas where the model fails to identify the modified regions correctly are marked in \textcolor{blue}{blue}. }
    \label{fig:example13}
\end{figure}

\begin{figure}[h]
    \centering
    \includegraphics[width=0.85\linewidth]{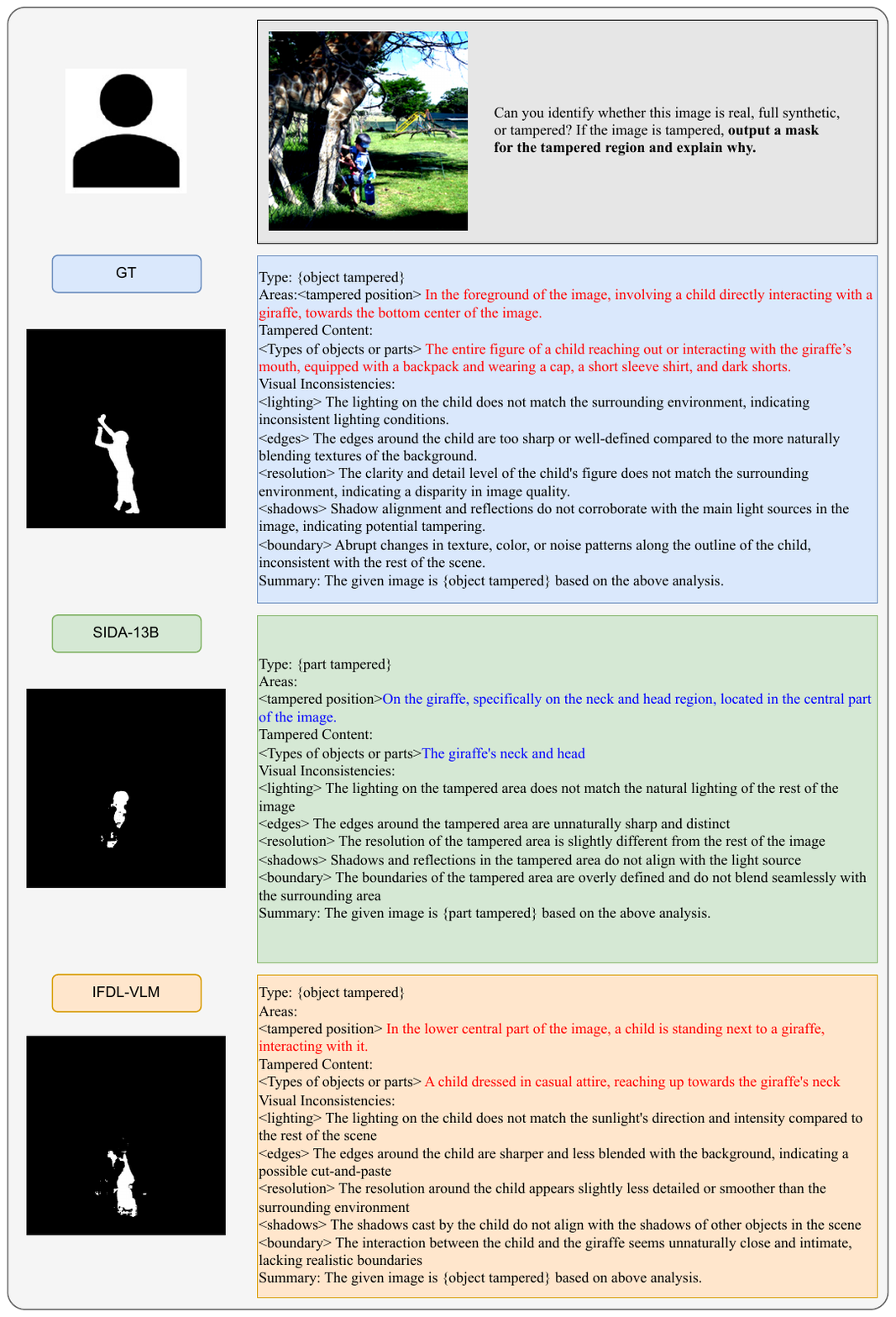}
    \caption{Qualitative comparison example. We highlight the accurate descriptions from GT in \textcolor{red}{red}, while the areas where the model fails to identify the modified regions correctly are marked in \textcolor{blue}{blue}. }
    \label{fig:example14}
\end{figure}

\begin{figure}[h]
    \centering
    \includegraphics[width=0.85\linewidth]{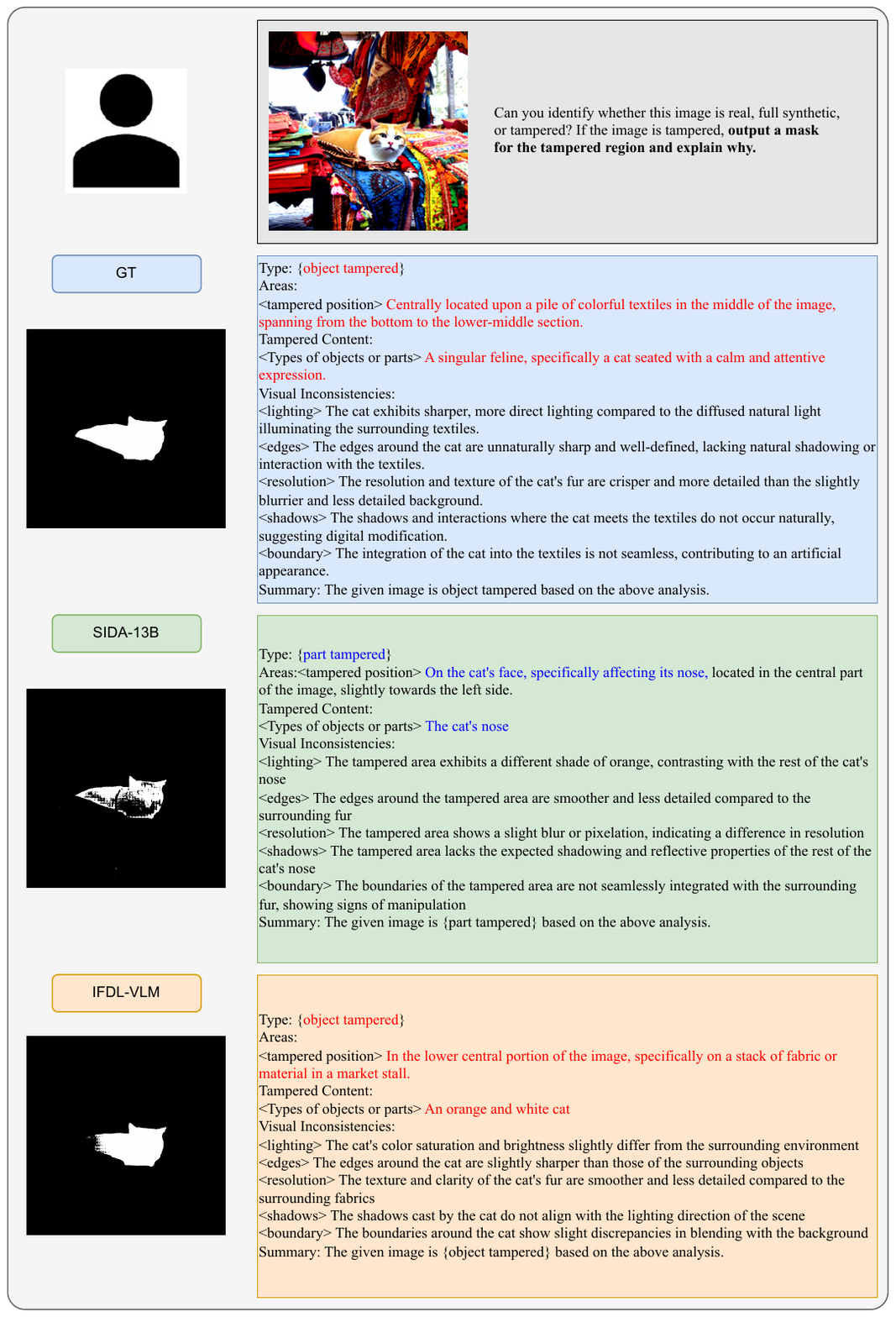}
    \caption{Qualitative comparison example. We highlight the accurate descriptions from GT in \textcolor{red}{red}, while the areas where the model fails to identify the modified regions correctly are marked in \textcolor{blue}{blue}. }
    \label{fig:example15}
\end{figure}

To demonstrate the advantages of our method, we show more qualitative comparison examples in Figures~\ref{fig:example1}, \ref{fig:example2}, \ref{fig:example3}, \ref{fig:example4}, \ref{fig:example5}, \ref{fig:example6}, \ref{fig:example7}, \ref{fig:example8}, \ref{fig:example9}, \ref{fig:example10}, \ref{fig:example11}, \ref{fig:example12}, \ref{fig:example13}, \ref{fig:example14}, \ref{fig:example15}.

\end{document}